\documentclass[pmlr]{template/jmlr}
\RequirePackage{graphicx}
 \usepackage{booktabs}
 \usepackage{multirow}
\usepackage{longtable}
\makeatletter
\def\set@curr@file#1{\def\@curr@file{#1}} 
\makeatother
\usepackage[load-configurations=version-1]{siunitx} 

\theorembodyfont{\upshape}
\theoremheaderfont{\scshape}
\theorempostheader{:}
\theoremsep{\newline}

 \jmlrproceedings{PMLR}{Proceedings of Machine Learning Research}
\jmlrvolume{340}
\jmlryear{2026}
\jmlrworkshop{Machine Learning for Healthcare}

\title[Learning Treatment Policies From Multimodal EHRs]{Annotation-Assisted Learning of Treatment Policies From Multimodal Electronic Health Records}

\author{\Name{Henri Arno}
       \Email{henri.arno@ugent.be}\\ 
       \addr Ghent University - imec, Ghent, Belgium
       \AND
       \Name{Thomas Demeester}
       \Email{thomas.demeester@ugent.be}\\ 
       \addr Ghent University - imec, Ghent, Belgium}

\begin{document}

\maketitle

\vspace{-0.8cm}
\begin{abstract}
We study how to learn treatment policies from multimodal electronic health records (EHRs) that consist of tabular data and clinical text. These policies can help physicians make better treatment decisions and allocate healthcare resources more efficiently.
Causal policy learning methods prioritize patients with the largest expected treatment benefit. Yet, existing estimators are designed for tabular covariates under causal assumptions that may be hard to justify in the multimodal setting. A pragmatic alternative is to apply causal estimators directly to multimodal representations, but this can produce biased treatment effect estimates when the representations do not preserve the relevant confounding information. As a result, predictive models of baseline risk are commonly used in practice to guide treatment decisions, although they are not designed to identify which patients benefit most from treatment.
We propose \textsc{AACE} (Annotation-Assisted Coarsened Effects), an \textit{annotation-assisted} approach to causal policy learning for multimodal EHRs. The method uses expert-provided annotations during training to support confounding adjustment, and then predicts treatment benefit from only multimodal representations at inference.
We show that the proposed method achieves strong empirical performance across synthetic, semi-synthetic, and real-world EHR datasets, outperforming risk-based and representation-based causal baselines, and offering practical insights for applying causal machine learning in clinical practice.
\end{abstract}

\section{Introduction}
\label{sec:intro}

Machine learning (ML) is increasingly being used in healthcare to support clinical decision-making \citep{esteva2019guide, rajkomar2019machine}. A central problem in this context is learning from observational data which patients should receive treatment, supporting more targeted and efficient allocation of clinical resources \citep{athey2021policy, nie2021learning}. Electronic health records (EHRs) are widely available in modern healthcare systems and capture rich, patient-specific information, making them well suited for learning data-driven treatment policies \citep{tang2024harnessing, shickel2017deep}. However, these records are typically \textbf{multimodal}, combining tabular data and clinical text. In this work, we study how to learn \textit{effective} and \textit{reliable} treatment policies from such multimodal EHR data.

One strategy to tackle this problem is to directly estimate treatment effects and prioritize patients with the largest expected benefit \citep{naturepaper, bica2021real}. By modeling how outcomes change under different treatment assignments, this strategy aligns the learning objective with the goal of making well-targeted treatment decisions. However, existing estimators are designed for \textit{tabular} covariates under causal identification assumptions, such as unconfoundedness. These assumptions can be difficult to justify in \textit{multimodal} EHRs, where patient information is distributed across tabular data and clinical text. A pragmatic alternative is therefore to apply causal estimators directly to representations of the multimodal data, implicitly assuming that they are sufficient for confounding adjustment. However, when relevant confounding information is missing from these representations, this leads to \textbf{biased} treatment effect estimates.

Because of these challenges, a more common strategy in practice is \textit{predictive} modeling of the clinical outcome in absence of treatment \citep{van2021clinical, goldstein2016opportunities}. The resulting predictions form an estimate of the patients' so-called \emph{baseline risk} and treatment can be prioritized for those most at risk. This approach is appealing because predictive models can easily handle multimodal data, and richer data often translates into higher predictive accuracy. However, as these models do not target treatment effects directly, they may fail to identify the patients who would benefit most, especially when baseline risk and treatment benefit are not well aligned. For instance, a patient may be at high baseline risk yet benefit little from treatment, while another with lower risk may benefit more, leading to \textbf{suboptimal} treatment allocation.

\paragraph{Proposed Method.}
We extend the causal strategy to the multimodal setting with an \textit{annotation-assisted} approach, which we refer to as \textsc{AACE} (Annotation-Assisted Coarsened Effects). The method builds on the doubly robust learner~\citep{kennedy2023towards} to estimate \emph{coarsened} treatment effects, which quantify the expected treatment benefit given the information contained in the multimodal representations. When relevant confounders are expressed in clinical text, but not entirely preserved in these representations, reliable treatment effect estimation requires additional training-time covariates that are sufficient for confounding adjustment. In \textsc{AACE}, expert-provided annotations serve as a pragmatic way to obtain such covariates during training. At inference time, the model relies \textit{only} on the multimodal representations to assign treatment.

\paragraph{Main Results.}
Our experiments show that the proposed method can learn effective treatment policies from multimodal EHR data given suitable supervision. On synthetic and semi-synthetic benchmarks, \textit{\textsc{AACE} outperforms both risk-based and representation-based causal baselines}. Additional robustness analyses show that the method remains effective under moderately imperfect annotations and representations. On real-world clinical data, the different methods produce substantially different treatment assignments, showing that the choice of method directly affects which patients are prioritized for treatment.

\paragraph{Contributions.}
Together, these findings motivate the following three contributions:

\begin{itemize}
\item We study the problem of learning treatment policies from multimodal EHR data, where patient information is distributed across tabular data and clinical text, and standard causal assumptions may not hold for the data representations directly.

\item We introduce \textsc{AACE}, an \textit{annotation-assisted} method for causal policy learning on multimodal data, which uses expert annotations during training to support confounding adjustment while relying only on multimodal representations at inference.

\newpage
\item We empirically evaluate the proposed method across synthetic, semi-synthetic, and real-world EHR datasets. The results show that \textsc{AACE} (i) outperforms both risk-based and representation-based causal baselines, (ii) remains effective under moderately imperfect annotations and representations, and (iii) generates materially different treatment assignments on real-world data.

\end{itemize}

\subsection*{Generalizable Insights about Machine Learning in the Context of Healthcare}
Our findings suggest several broader lessons for machine learning in healthcare. In multimodal settings, predictive models of baseline risk are not necessarily appropriate for treatment assignment, because risk and treatment benefit do not necessarily align. At the same time, directly applying causal estimators to learned representations can be problematic when those representations do not capture all relevant confounders, leading to biased effect estimates. Expert-provided annotations offer a \textit{practical} way to support confounding adjustment during training, helping causal methods better match the realities of multimodal clinical data. More broadly, the choice of modeling strategy can directly affect which patients are prioritized for treatment, making the method choice matter in practice.

\section{Background and Related Work}
\label{sec:related}

We summarize the foundations of learning treatment policies from observational data and review related work on the main strategies for this task.

\subsection{Learning Treatment Policies From Observational Data}
Consider a setting where each individual~$i$ in the population is characterized by covariates~$X_i \in \mathcal{X}$, receives a binary treatment~$T_i \in \{0,1\}$, and has an associated continuous outcome~$Y_i \in \mathbb{R}$. In the Rubin-Neyman potential outcomes framework~\citep{rubinneyman}, each individual also has two potential outcomes,~$Y_i(1)$ and~$Y_i(0)$, where~$Y_i(t)$ denotes the outcome that we would observe if individual~$i$ were assigned treatment~$T_i = t$. In practice, only one of the two potential outcomes is observed, corresponding to the treatment that was actually received.

A \emph{treatment policy}~$\pi$ is a rule that maps individual covariates to a treatment decision $\pi(x): \mathcal{X}\!\to\!\{0,1\}$. In the potential outcomes framework, the performance of a policy can be quantified by its \emph{policy value},
\begin{equation}
V(\pi) = \mathbb{E}\big[\,Y\big(\pi(X)\big)\big]
       = \mathbb{E}\big[\,\pi(X)\,Y(1) + \big(1-\pi(X)\big)\,Y(0)\big],
\end{equation}
which represents the mean outcome in the population if treatment were assigned according to~$\pi$. The optimal policy~$\pi^*$ is the one that maximizes the policy value (assuming larger outcomes are better). In many applications, however, treatment resources are limited and only $k$ individuals in a population \emph{can} be treated. In such constrained settings, the optimal policy maximizes the policy value while satisfying the resource constraint. 

In practice, we aim to learn treatment policies from observed data $(X, T, Y) \sim P$, where the joint distribution factorizes as $P(X, T, Y)=P(X)\,P(T \mid X)\,P(Y \mid X, T)$. Here, $P(T \mid X)$ represents the existing, possibly suboptimal, \emph{observational} policy $\pi_{\mathrm{obs}}(x)$. The goal of treatment policy learning is to use this data to estimate a new policy~$\hat\pi$ that achieves a higher policy value than~$\pi_{\mathrm{obs}}$ and approximates~$\pi^*$.

\subsection{Causal Treatment Policies}
\label{sec:causalstrat}

Causal strategies learn treatment policies by modeling how outcomes would change under different treatment assignments. Within this framework, two main approaches have emerged. First, in \emph{direct policy learning}, the treatment policy is learned directly by optimizing an empirical estimate of the policy value~\citep{qian2011performance, kallus2018balanced, athey2021policy}. Second, in \emph{CATE-based policy learning}, the conditional average treatment effect (CATE) is estimated first, and individuals whose estimated benefit exceeds a threshold are treated~\citep{ranker, frauendecision, loria1}. We focus on this latter family because it forms the basis of our proposed method.

The conditional average treatment effect (CATE) quantifies the expected benefit of treatment for an individual with covariates $X=x$,
\begin{equation}
    \tau^x(x) = \mathbb{E}\big[Y(1)-Y(0)\mid X=x\big].
\end{equation}
Because the potential outcomes are not directly observable, we rely on the following standard causal inference assumptions for the identification of the CATE:
\begin{itemize}
\item \textbf{Assumption 1.1 (consistency).} The observed outcome equals the potential outcome under the received treatment: $Y(t)=Y$ whenever $T=t$.
\item \textbf{Assumption 1.2 (positivity).} Each individual has a non-zero probability of receiving either treatment: $0 < e(x) < 1$, where $e(x)=\pi_{\mathrm{obs}}(x)=P(T=1\mid X=x)$ is the \emph{propensity score}.
\item \textbf{Assumption 1.3 (unconfoundedness).} There are no unmeasured confounders: $Y(t)\perp\!\!\!\perp T \mid X$.
\end{itemize}

Under these assumptions, the CATE is identified, meaning that it can be estimated from observational data, and may be expressed as
\begin{equation}
    \tau^x(x) = \mathbb{E}\big[Y\mid T{=}1, X{=}x\big] - \mathbb{E}\big[Y\mid T{=}0, X{=}x\big] = \mu_1(x)-\mu_0(x),
\end{equation}
where $\mu_t(x)=\mathbb{E}\big[Y\mid T=t, X=x\big]$ denotes the conditional mean outcome under treatment $t\in\{0,1\}$, commonly referred to as the \emph{response surfaces}. Together with the propensity score, these quantities form the \emph{nuisance functions}, collectively referred to as $\eta = (e, \mu_1, \mu_0)$.

There is a rich literature on estimating the CATE from observational data. Existing machine learning models have been adapted for this purpose, including Gaussian processes, random forests, and generative adversarial networks~\citep{gausian_processes, wager2018estimation, yoon2018ganite}. More recently, meta-learners have gained popularity as general-purpose procedures for CATE estimation that can be instantiated with standard supervised learning models~\citep{kunzel2019metalearners, nie2021quasi, kennedy2023towards}. In their standard form, however, these estimators are typically developed for \textit{tabular} covariates that are directly observed, and satisfy the above assumptions (1.1 -- 1.3). In \textit{multimodal} clinical data, these assumptions are difficult to satisfy in practice, because relevant confounding information may be distributed across tabular variables and clinical text, or may be only imperfectly preserved in the data representations.

\subsection{Representation-Based CATE Estimation}
Several authors have proposed neural network architectures that learn intermediate representations of \textit{tabular} covariates for CATE estimation under the standard causal inference assumptions. Neural architectures such as TARNet learn a shared representation of the covariates and use treatment-specific heads to estimate the potential outcomes \citep{johansson2016learning, shalit3}. DragonNet extends this design with an additional head for the propensity score, while subsequent work has explored broader architectural variants and inductive biases for representation-based treatment effect estimation \citep{shi2019adapting, curth2021inductive}. However, when these learned representations fail to preserve relevant confounding information, the resulting treatment effect estimates can be biased \citep{melnychuk2023bounds}.

Recent work has also studied treatment effect estimation from high-dimensional and multimodal data, particularly text \citep{llmdriventreatment, egami2022make, daoud-etal-2022-conceptualizing, gultchin2021operationalizing, veitch2020adapting, keith-etal-2020-text, wood2018challenges}, and more recently images, and combinations of both \citep{zhu2025optimizing, klaassen2024doublemldeep, jerzak23a, pmlr-v177-sanchez22a}. In these settings, the learned representations are typically assumed to be \textit{sufficient for confounding adjustment}. In contrast, we argue that this assumption can be hard to satisfy in multimodal EHRs, and instead propose \textit{annotation-assisted} learning as a more practical alternative. This is closely related to learning with privileged information, where richer variables are available during training than at inference~\citep{lopez2015unifying,makar2019distillation}. \textsc{AACE} applies this idea to multimodal confounding, using expert annotations for confounding adjustment before learning treatment effects over inference-time representations.

\subsection{Risk-Based Treatment Policies}
An alternative strategy for learning treatment policies is to rely on \emph{predictive} or \emph{risk-based} models that estimate an individual’s expected outcome in the absence of treatment, $\mu_0(x)=\mathbb{E}\big[Y\mid T=0, X=x\big]$. This quantity can be interpreted as an individual’s baseline risk, and treatment may then be prioritized for those with the highest predicted risk. Recent work has compared such strategies to causal methods \citep{loria1, loria3, nudge}, showing that risk-based policies \emph{can} perform competitively, for instance when treatment effects and baseline risk are strongly correlated, or when treatment effects are difficult to estimate in small samples.

This strategy is particularly appealing in the multimodal setting, because it does not rely on causal identification assumptions, and predictive accuracy often improves with richer data. For instance, clinical outcome prediction can improve substantially when incorporating the multiple modalities in electronic health records \citep{liu2018deep, yang-wu-2021-leverage}. However, while such models may predict outcomes well, our goal is to learn \emph{effective treatment policies} rather than to optimize predictive accuracy alone.

\section{Problem Setting}
\label{sec:problem}
In this section, we describe the structure of the multimodal electronic health records considered in this work and how expert annotations can address the confounding challenges they introduce.

\subsection{Multimodal Confounding}

\begin{figure}[t]
    \vspace{-0.4cm}
    \centering
    \includegraphics[width=0.8\columnwidth]{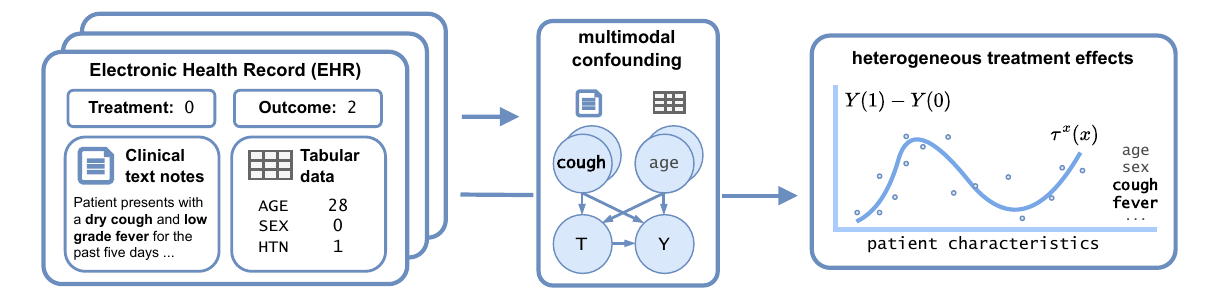}
    \vspace{-0.3cm}
    \caption{Overview of the problem setting. \textbf{Left:} Electronic health records combine tabular data and clinical text: key confounders (e.g., \textit{cough} and \textit{fever}) may appear only in text. \textbf{Middle:} Multimodal confounding influences both treatment \(T\) and outcome \(Y\). \textbf{Right:} Heterogeneity in the conditional average treatment effect \(\tau^x(x)\) across patient characteristics.}
    \label{fig:problem}
    \vspace{-0.2cm}
\end{figure}

Electronic health records (EHRs) are routinely collected in healthcare systems and capture detailed patient information, making them a natural choice for learning treatment policies from real-world data. These records typically combine tabular variables, such as demographics and lab measurements, with clinical text, such as nursing notes describing patient symptoms or radiology reports summarizing imaging findings \citep{johnson2016mimic}. Because of this multimodality, clinical factors that influence both treatment decisions and patient outcomes may appear only in the text, as illustrated in Figure~\ref{fig:problem}. Estimating treatment effects from such data therefore requires adjustment for confounders that may be distributed across modalities.

\subsection{Annotation-Assisted Confounding Adjustment}

The clinical notes in an EHR can be encoded using a pre-trained text model, such as ModernBERT~\citep{modernbert}, to obtain an embedding. These embeddings, once concatenated with the tabular variables, form a multimodal representation $\phi$ of a record. However, if the text captures confounding factors only partially, or if the encoder does not preserve this information entirely, \textit{unconfoundedness does not hold with respect to $\phi$.} In that case, treatment effects are \textbf{not} identified from observational data $(\phi, T, Y)$ alone.

Reliable treatment effect estimation in this setting therefore requires training-time covariates that are sufficient for confounding adjustment. In multimodal EHRs, when clinically relevant confounders are expressed in the text, but not reliably preserved in $\phi$, some explicit representation of these variables is \textbf{necessary} during training. In this work, we argue that \textit{expert-provided annotations} offer a pragmatic way to obtain such covariates. Concretely, the annotations should capture clinically relevant factors, documented in the notes, that influence both treatment decisions and patient outcomes. In the example of Figure~\ref{fig:problem}, this would correspond to annotating symptom variables such as \emph{cough} or \emph{fever}. 

Which covariates need to be annotated is application-dependent and should be guided by domain expertise. We therefore deliberately do not prescribe a fixed annotation protocol. In practice, such annotations may be obtained through a one-time annotation effort on top of routine clinical documentation, or through retrospective expert review. Appendix~\ref{app:annotations} provides a concrete example of how such annotations may be obtained in practice. 

While this introduces an additional annotation cost, we view it as a practical compromise: rather than assuming that the representation alone preserves all confounding information, we explicitly collect the variables needed to support reliable confounding adjustment. Formally, we let $X$ denote the training-time covariates that are assumed sufficient to adjust for confounding, i.e., $Y(t) \!\perp\!\!\!\perp T \mid X$, together with consistency and positivity. In our setting, $X$ comprises the tabular variables in the EHR, together with the expert-provided annotations of confounders expressed in the text. The training data therefore consists of $(\phi, X, T, Y)$, whereas at inference, only the representation $\phi$ is available.

\section{Proposed Method: \textsc{AACE}}
\label{sec:method}
We now describe \textsc{AACE} (Annotation-Assisted Coarsened Effects), our \textit{annotation-assisted} method for treatment policy learning from multimodal EHR data. The method uses expert-provided annotations to support confounding adjustment during training, and then learns to predict treatment benefit from multimodal representations alone at inference time.

\begin{figure}[t]
    \vspace{-0.3cm}
    \centering
    \includegraphics[width=\columnwidth]{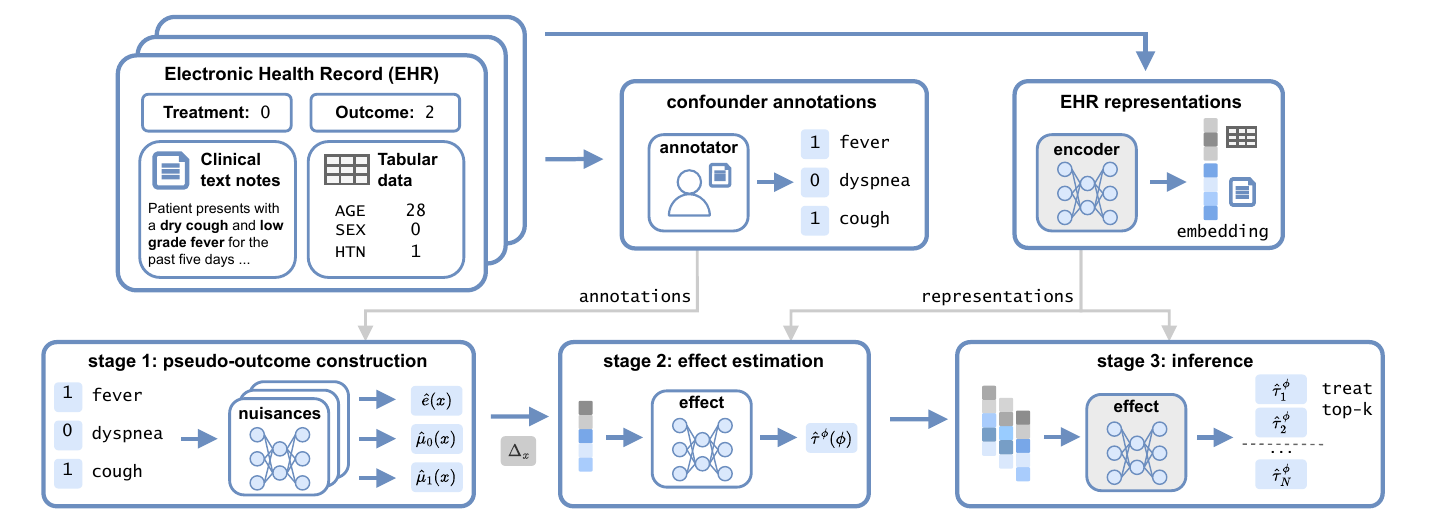}
    \caption{Overview of \textsc{AACE}. \textbf{Top:} Each multimodal electronic health record is (1) annotated during training to capture confounders and (2) encoded into a multimodal representation using a pre-trained model. \textbf{Bottom:} \textsc{AACE} proceeds in three stages: (1) nuisance estimation and pseudo-outcome construction using the causally sufficient training covariates, (2) treatment effect estimation from multimodal representations, and (3) inference-time treatment assignment.}
    \label{fig:method}
\end{figure}

\subsection{Coarsened Treatment Effects.}

Because treatment decisions at inference can only depend on the multimodal representations, we define the target estimand as the expected treatment effect conditional on $\phi$,
\begin{equation}
    \label{eq:targetestimand}
    \tau^{\phi}(\phi) 
    := \mathbb{E}\big[Y(1)-Y(0) \mid \phi\big] = \int \tau^x(x)\, p(x \mid \phi)\, dx .
\end{equation}
Thus, $\tau^\phi(\phi)$ averages the treatment effects $\tau^x(x)$ over the patient subgroups whose covariates are compatible with the representation. It interpolates between the average treatment effect (ATE) when $\phi$ is uninformative about $X$, and the conditional average treatment effect (CATE) when $\phi$ fully determines $X$. The equality in~\eqref{eq:targetestimand} follows from the law of iterated expectations, under the assumption that the representations provide no additional information about the potential outcomes beyond the covariates, i.e., $Y(t) \perp\!\!\!\perp \phi \mid X$ for $t \in \{0,1\}$. In our setting, this is natural because $\phi$ is derived from clinical data that encode the underlying covariates.

\paragraph{Treatment Policies Based on Coarsened Effects.}
As discussed in Section~\ref{sec:causalstrat}, CATE-based policies allocate treatment to the individuals with the highest expected treatment benefit. When the multimodal representations lose confounding information, the coarsened effects $\tau^\phi(\phi)$ may differ from the true effects $\tau^x(x)$. Nonetheless, the implied treatment policy can remain \textit{optimal} when the bias induced by coarsening is small relative to the treatment effect margins. To see this, let $\delta_i = | \tau^x(x_i) - \tau^\phi(\phi_i) |$ denote the coarsening bias for individual $i$. For a fixed treatment budget $k$, let $\gamma_k$ denote the treatment effect margin at the decision threshold. If individuals are ordered by decreasing (true) treatment effects, then $\gamma_k$ is the difference between the $k$-th and $(k+1)$-th largest treatment effects.

\begin{proposition}
\label{prop:topk}
For a fixed treatment budget $k$, if $\max_i \delta_i < \frac{\gamma_k}{2}$, then ranking individuals by $\tau^\phi(\phi)$ yields the same top-$k$ treatment assignment as ranking them by $\tau^x(x)$.
\end{proposition}

Thus, representation-based coarsening can distort effect values without affecting treatment assignment, as long as no individuals are swapped across the treatment decision boundary. A proof is given in Appendix~\ref{app:ordering}. In practice, the condition of Proposition~\ref{prop:topk} cannot be verified directly, since the true treatment effects are unknown, much like the standard causal identification assumptions cannot be verified from observational data alone (Assumptions 1.1 -- 1.3). However, if the annotated covariates can be recovered reasonably well from the representations in the training data, this is a strong indication that the coarsening bias is limited and unlikely to alter treatment decisions.

\subsection{Estimation with Annotation-Assisted Learning.}
To estimate the coarsened treatment effects, \textsc{AACE} proceeds in three stages (cf. Figure~\ref{fig:method}).

\paragraph{Stage 1: Pseudo-Outcome Construction.}
We begin by constructing pseudo-outcomes following the doubly robust (DR) learner, a standard meta-learner for treatment effect estimation~\citep{kennedy2023towards}. This step uses the additional training-time covariates $X$ to estimate the nuisance functions: the propensity score $\hat e(x)$ and the response models $\hat\mu_t(x)$ for $t \in \{0,1\}$. The resulting pseudo-outcome is
\begin{equation}
\Delta_x
=
\hat\mu_1(x) - \hat\mu_0(x)
+
\frac{T - \hat e(x)}{\hat e(x)\big(1-\hat e(x)\big)}\big(Y - \hat\mu_T(x)\big).
\end{equation}
By the doubly robust property, if either the propensity model or the response models are correctly specified, then $\mathbb{E}[\Delta_x \mid X=x] = \tau^x(x).$ Thus, $\Delta_x$ provides an unbiased signal for the treatment effect, given the causally sufficient training-time covariates $X$.

\paragraph{Stage 2: Effect Estimation From Multimodal Representations.}
In the second stage, we regress the pseudo-outcomes $\Delta_x$ on the multimodal representations $\phi$. By the law of iterated expectations,
\begin{equation}
\mathbb{E}\big[\Delta_x \mid \phi\big]
=
\mathbb{E}\!\left[\mathbb{E}\big[\Delta_x \mid X\big] \mid \phi\right]
=
\mathbb{E}\big[\tau^x(X) \mid \phi\big]
=
\tau^\phi(\phi),
\end{equation}
so this stage directly learns the target estimand introduced in~\eqref{eq:targetestimand}. In other words, the model learns how treatment benefit varies with the information in the multimodal representations.

\paragraph{Stage 3: Inference-Time Treatment Assignment.}
At inference time, only the multimodal representation $\phi$ is required. Each EHR is first encoded into its representation, after which the trained second-stage model predicts the corresponding coarsened treatment effect. Treatment is then assigned to the top-$k$ individuals with the highest predicted benefit.

\section{Experiments}
We describe the experimental setup and then evaluate \textsc{AACE} on synthetic, semi-synthetic, and real-world multimodal EHR benchmarks. Full details are provided in Appendices~\ref{app:datasets}--\ref{app:details}.\footnote{Code available at \url{https://github.com/henriarnoUG/causal-ehr}.}

\subsection{Datasets}

\paragraph{SynSum.} The fully synthetic \textsc{SynSum} dataset consists of 10{,}000 medical records with tabular variables and free-text clinical notes \citep{SynSUM}. The notes are generated with an LLM and describe patient symptoms such as \textit{fever}, \textit{cough}, and \textit{pain}, which act as text-based confounders and influence both treatment and outcome. We generate the treatments and outcomes such that confounding is substantial and baseline risk is only moderately aligned with treatment benefit. Expert annotations are simulated using the true symptom values used to generate the text, and we additionally study how performance changes when these annotations are imperfect. This results in a controlled multimodal EHR benchmark with known ground-truth treatment effects.

\paragraph{MIMIC-Syn.} The semi-synthetic \textsc{MIMIC-Syn} benchmark is derived from the publicly available \textsc{MIMIC-III} database of intensive care EHRs~\citep{johnson2016mimic}. Following~\citet{chen2024proximal}, we retain the real clinical notes and demographic variables to construct the multimodal EHRs. We then generate treatments and outcomes using cardiovascular diagnoses, age, and sex. The diagnosis indicators act as text-based confounders and influence both treatment and outcome, while being only implicitly expressed in the clinical text. Expert annotations are simulated using these diagnosis indicators. This results in a semi-synthetic multimodal EHR benchmark with known ground-truth treatment effects and realistic covariate distributions.

\paragraph{MIMIC-Real.} The \textsc{MIMIC-Real} dataset was also derived from \textsc{MIMIC-III}~\citep{johnson2016mimic}. Each multimodal EHR consists of demographic variables and clinical notes recorded prior to treatment. The treatment corresponds to \textit{antibiotic administration} during ICU admission, and the outcome is \textit{in-hospital mortality}. Vital-sign measurements are used as additional training-time covariates, as they plausibly influence both treatment and outcomes. Since this is an observational dataset, the required unconfoundedness assumption cannot be verified, and ground-truth effects are unknown. We therefore \textbf{cannot} use \textsc{MIMIC-Real} to evaluate which policy is best, and instead use it \textit{only} to qualitatively compare the treatment assignments of the different methods in a real-world setting. The goal of this experiment is therefore to study \textbf{how treatment decisions differ} in practice, rather than to determine which policy is optimal.

\subsection{Baselines and Evaluation}

\paragraph{Baselines.}
We compare \textsc{AACE} to two standard strategies for treatment policy learning from multimodal EHR data. First, we consider \textit{risk-based models}, which prioritize patients based on predicted baseline risk. Second, we consider \textit{representation-based causal models}, which estimate treatment effects directly from the multimodal representations, implicitly assuming that these representations are sufficient for confounding adjustment. For this class, we apply standard neural CATE estimators, namely TARNet~\citep{shalit3}, DragonNet~\citep{shi2019adapting}, and DR-learner~\citep{kennedy2023towards}, directly to the multimodal representations. All methods use the same frozen ModernBERT-based multimodal representations as input for fair comparison.

\paragraph{Evaluation.}
For datasets with known ground-truth effects (\textsc{SynSum} and \textsc{MIMIC-Syn}), we evaluate the treatment policies induced by each method under varying treatment budgets. Each method produces a ranking of individuals, either by predicted treatment benefit or by baseline risk, and we evaluate the policy obtained by treating the top-$k$ individuals. We report four metrics: AUTOC (area under the treatment-outcome curve)~\citep{yadlowsky2025evaluating}, which measures ranking quality across treatment budgets; the mean policy value across all budgets $\big(\overline{V}\big)$; the policy value with a treatment budget of $10\%$ of the population ($V_{10\%}$); and PEHE (precision in estimating heterogeneous effects), which measures treatment effect estimation error. Lower values for all metrics indicate better performance.

\subsection{Results on SynSum}

The \textsc{SynSum} benchmark provides a controlled setting with substantial confounding and moderate alignment between baseline risk and treatment benefit. Table~\ref{tab:synsum_policy} reports the policy evaluation results across training sizes. \textsc{AACE} consistently outperforms both risk-based and representation-based causal baselines on all metrics and at all training sizes. In particular, methods that estimate treatment effects directly from the multimodal representations (DR-Emb, DragonNet, TARNet) underperform compared to \textsc{AACE}, indicating that the additional \textit{targeted} supervision is beneficial for confounding adjustment on this benchmark. Additional results in Appendix~\ref{app:results} show that the confounders are partially recoverable from the text representations, and report the policy metrics for additional training sizes.

\begin{table*}[t]
\centering
\caption{\textbf{\textsc{SynSum} (main results).} Policy metrics across training sizes, evaluated on a fixed test set (mean~$\pm$~std.~dev. over five random seeds). Lower values are better. PEHE measures treatment effect estimation error and is not defined for the risk-based model. The best model within each training size is shown in \textbf{bold}. The \textit{oracle} row reports the metrics of a policy that ranks by the true treatment effect.}
\label{tab:synsum_policy}
\renewcommand{\arraystretch}{1.1}
\begin{scriptsize}
\begin{tabular}{lccccc}
\toprule
\textbf{Training size} & \textbf{Model} & \textbf{AUTOC $(\downarrow)$} & \textbf{$\boldsymbol{\overline{V}}$ $(\downarrow)$} & \textbf{$\boldsymbol{V_{10\%}}$ $(\downarrow)$} & \textbf{PEHE $(\downarrow)$} \\
\midrule
\multirow{5}{*}{1{,}000}
 & Risk & $-0.25 \pm 0.02$ & $0.77 \pm 0.01$ & $1.17 \pm 0.01$ & --- \\
 & DR-Emb & $-0.42 \pm 0.04$ & $0.69 \pm 0.01$ & $1.14 \pm 0.01$ & $0.71 \pm 0.02$ \\
 & DragonNet & $-0.09 \pm 0.16$ & $0.79 \pm 0.04$ & $1.21 \pm 0.04$ & $0.96 \pm 0.02$ \\
 & TARNet & $-0.43 \pm 0.03$ & $0.69 \pm 0.01$ & $1.14 \pm 0.01$ & $0.70 \pm 0.03$ \\
 & AACE (ours) & $\mathbf{-0.49 \pm 0.02}$ & $\mathbf{0.68 \pm 0.01}$ & $\mathbf{1.12 \pm 0.01}$ & $\mathbf{0.62 \pm 0.02}$ \\
\midrule
\multirow{5}{*}{4{,}000}
 & Risk & $-0.22 \pm 0.01$ & $0.77 \pm 0.00$ & $1.18 \pm 0.00$ & --- \\
 & DR-Emb & $-0.51 \pm 0.01$ & $0.66 \pm 0.00$ & $1.12 \pm 0.00$ & $0.59 \pm 0.02$ \\
 & DragonNet & $-0.40 \pm 0.02$ & $0.70 \pm 0.00$ & $1.14 \pm 0.01$ & $0.71 \pm 0.02$ \\
 & TARNet & $-0.51 \pm 0.02$ & $0.66 \pm 0.00$ & $1.13 \pm 0.01$ & $0.58 \pm 0.02$ \\
 & AACE (ours) & $\mathbf{-0.56 \pm 0.01}$ & $\mathbf{0.65 \pm 0.00}$ & $\mathbf{1.11 \pm 0.01}$ & $\mathbf{0.53 \pm 0.02}$ \\
\midrule
\multirow{5}{*}{8{,}000}
 & Risk & $-0.23 \pm 0.00$ & $0.77 \pm 0.00$ & $1.18 \pm 0.01$ & --- \\
 & DR-Emb & $-0.53 \pm 0.01$ & $0.65 \pm 0.00$ & $1.12 \pm 0.00$ & $0.56 \pm 0.01$ \\
 & DragonNet & $-0.48 \pm 0.01$ & $0.67 \pm 0.00$ & $1.13 \pm 0.00$ & $0.63 \pm 0.02$ \\
 & TARNet & $-0.53 \pm 0.00$ & $0.65 \pm 0.00$ & $1.12 \pm 0.00$ & $0.58 \pm 0.02$ \\
 & AACE (ours) & $\mathbf{-0.58 \pm 0.00}$ & $\mathbf{0.64 \pm 0.00}$ & $\mathbf{1.11 \pm 0.00}$ & $\mathbf{0.50 \pm 0.01}$ \\
\midrule
\multicolumn{2}{l}{Oracle} & $-0.74$ & $0.61$ & $1.08$ & 0.00 \\
\bottomrule
\end{tabular}
\end{scriptsize}
\end{table*}

\subsection{Results on MIMIC-Syn}
Table~\ref{tab:mimic_syn} reports the results on the semi-synthetic \textsc{MIMIC-Syn} benchmark. \textsc{AACE} again achieves the strongest performance overall. It obtains the best AUTOC, $V_{10\%}$, and PEHE, while remaining highly competitive on the mean policy value, where DragonNet performs marginally better. These results show that the advantage of annotation-assisted learning is not limited to the synthetic setting, but persists in a more realistic benchmark with real multimodal covariate distributions. Additional results in Appendix~\ref{app:results} show that the confounders are partially recoverable from the clinical text representations.

\begin{table*}[h]
\centering
\caption{\textbf{\textsc{MIMIC-Syn} (main results).} Policy metrics on a fixed test set (mean~$\pm$~std.~dev. over twenty random seeds). See Table~\ref{tab:synsum_policy} for metric definitions and interpretation.}
\label{tab:mimic_syn}
\renewcommand{\arraystretch}{1.1}
\begin{scriptsize}
\begin{tabular}{lcccc}
\toprule
\textbf{Model} & \textbf{AUTOC $(\downarrow)$} & \textbf{$\boldsymbol{\overline{V}}$ $(\downarrow)$} & \textbf{$\boldsymbol{V_{10\%}}$ $(\downarrow)$} & \textbf{PEHE $(\downarrow)$} \\
\midrule
Risk & $0.03 \pm 0.00$ & $6.49 \pm 0.00$ & $6.96 \pm 0.00$ & --- \\
DR-Emb & $-0.12 \pm 0.06$ & $6.45 \pm 0.01$ & $6.92 \pm 0.02$ & $0.60 \pm 0.04$ \\
DragonNet & $-0.16 \pm 0.04$ & $\mathbf{6.44 \pm 0.01}$ & $6.91 \pm 0.01$ & $0.57 \pm 0.04$ \\
TARNet & $-0.16 \pm 0.04$ & $6.44 \pm 0.01$ & $6.91 \pm 0.01$ & $0.60 \pm 0.06$ \\
AACE (ours) & $\mathbf{-0.17 \pm 0.06}$ & $6.45 \pm 0.01$ & $\mathbf{6.90 \pm 0.02}$ & $\mathbf{0.45 \pm 0.01}$ \\
\midrule
Oracle & $-0.43$ & $6.37$ & $6.84$ & $0.00$ \\
\bottomrule
\end{tabular}
\end{scriptsize}
\end{table*}

\subsection{Robustness Analyses}

We further analyze how \textsc{AACE} depends on the quality of its annotations and multimodal representations. All analyses in this subsection are performed on the \textsc{SynSum} dataset, extended results are reported in Appendix~\ref{app:results}.

\begin{figure}[t]
    \centering
    \begin{minipage}[t]{0.46\textwidth}
        \centering
        \textbf{(a)} Second-stage input ablation\\[5pt]
        \includegraphics[width=\linewidth]{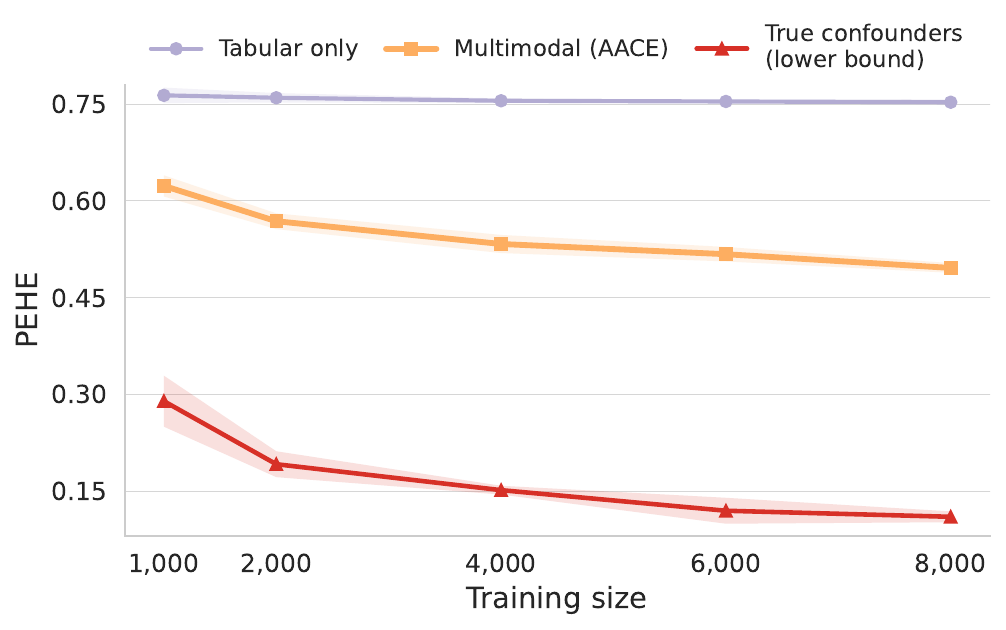}
    \end{minipage}
    \hfill
    \begin{minipage}[t]{0.52\textwidth}
        \centering
        \textbf{(b)} Multimodal representations\\[5pt]
        \includegraphics[width=\linewidth]{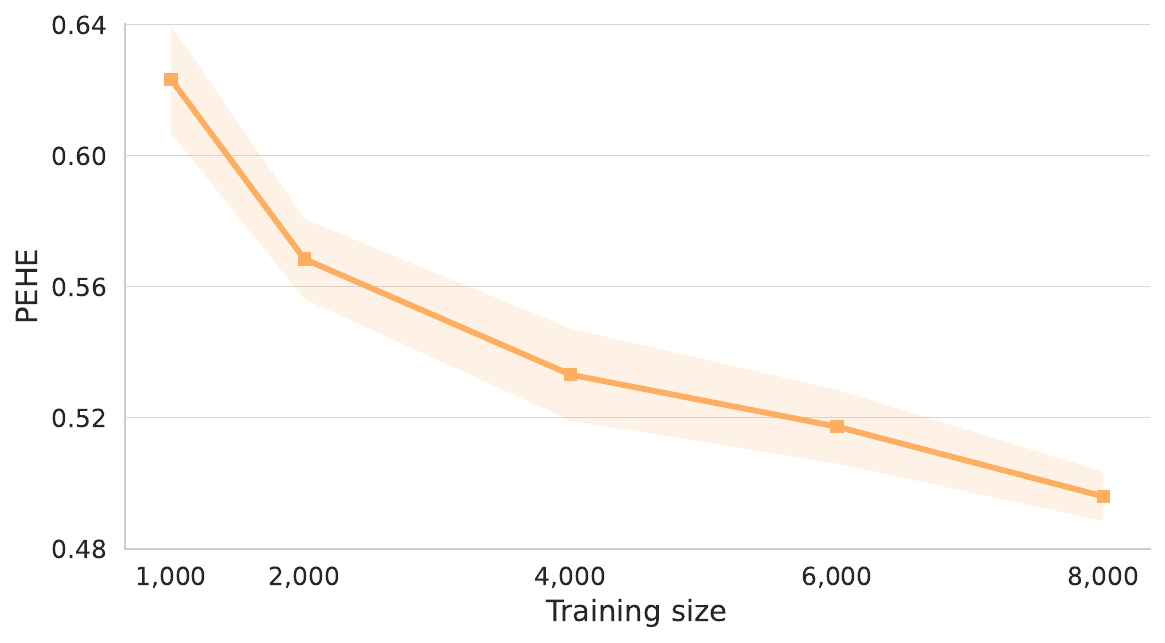}
    \end{minipage}
    \caption{\textbf{\textsc{SynSum} (learning coarsened treatment effects).} PEHE as a function of training size. Panel (a) reports a second-stage input ablation in which the same DR pseudo-outcomes are used throughout, while the effect model is trained using tabular variables only, multimodal representations, or the true confounders, which are not available in practice and therefore provide a lower bound on PEHE. Panel (b) shows a zoomed view of the multimodal setting used by \textsc{AACE}, where PEHE decreases steadily with training size. Lower values are better.}
    \label{fig:synsum_pehe_main}
    \vspace{-0.7cm}
\end{figure}

\paragraph{Input Ablation.}
Figure~\ref{fig:synsum_pehe_main} reports a second-stage input ablation in which the same DR pseudo-outcomes are used throughout, while the effect model is trained using either (i) only the tabular variables from the EHR, (ii) the multimodal representations used by \textsc{AACE}, or (iii) the true confounders, which are not available in practice and therefore provide a lower bound on PEHE. The multimodal representations substantially reduce treatment effect estimation error relative to tabular-only inputs, and the error decreases steadily as more training data becomes available. This suggests that the multimodal representations capture relevant confounding signal for learning the coarsened treatment effects.

\paragraph{Annotation Quality.}
We next assess the robustness of \textsc{AACE} to imperfect annotations. We consider three failure modes that may arise in practice: (i) noisy or incorrect annotations of relevant confounders, simulated by replacing annotations with random values drawn from the empirical covariate distribution, (ii) missing relevant confounders, simulated by removing annotated confounders before training, and (iii) irrelevant annotations, simulated by adding variables to the annotations that are unrelated to the true confounding structure. Tables~\ref{tab:synsum_annotation_corruption_main}--\ref{tab:synsum_irrelevant_annotations_main} show that noisy annotations and missing relevant confounders lead to gradual performance degradation, whereas adding irrelevant variables has little effect. Overall, these results indicate that \textsc{AACE} mainly depends on capturing the relevant confounding signal, while being relatively insensitive to irrelevant annotations.

\begin{table}[!h]
\centering
\caption{\textbf{\textsc{SynSum} (noisy annotations).} AUTOC of \textsc{AACE} as a function of annotation corruption, using 4{,}000 training samples and a fixed test set (mean~$\pm$~std.~dev. over five random seeds). Lower values are better.}
\label{tab:synsum_annotation_corruption_main}
\renewcommand{\arraystretch}{1.1}
\begin{scriptsize}
\begin{tabular}{l|cccccc}
\toprule
\textbf{Corr. (\%)} & 0 & 5 & 10 & 25 & 50 & 100 \\
\midrule
\textbf{AUTOC} & $-0.56 \pm 0.01$ & $-0.55 \pm 0.01$ & $-0.53 \pm 0.01$ & $-0.49 \pm 0.01$ & $-0.42 \pm 0.04$ & $-0.37 \pm 0.06$ \\
\bottomrule
\end{tabular}
\end{scriptsize}
\end{table}

\begin{table}[!h]
\centering
\caption{\textbf{\textsc{SynSum} (missing relevant confounders).} AUTOC of \textsc{AACE} as a function of the number of removed annotated confounders. See Table~\ref{tab:synsum_annotation_corruption_main} for details.}
\label{tab:synsum_dropped_annotations_main}
\renewcommand{\arraystretch}{1.1}
\begin{scriptsize}
\begin{tabular}{l|cccccc}
\toprule
\textbf{Dropped (n)} & 0 & 1 & 2 & 3 & 4 & 5 \\
\midrule
\textbf{AUTOC} & $-0.56 \pm 0.01$ & $-0.51 \pm 0.03$ & $-0.50 \pm 0.01$ & $-0.47 \pm 0.04$ & $-0.41 \pm 0.03$ & $-0.38 \pm 0.05$ \\
\bottomrule
\end{tabular}
\end{scriptsize}
\end{table}

\begin{table}[!h]
\centering
\caption{\textbf{\textsc{SynSum} (irrelevant annotations).} AUTOC of \textsc{AACE} as a function of the number of added irrelevant annotation variables. See Table~\ref{tab:synsum_annotation_corruption_main} for details.}
\label{tab:synsum_irrelevant_annotations_main}
\renewcommand{\arraystretch}{1.1}
\begin{scriptsize}
\begin{tabular}{l|cccccc}
\toprule
\textbf{Added (n)} & 0 & 1 & 2 & 3 & 4 & 5 \\
\midrule
\textbf{AUTOC} & $-0.56 \pm 0.01$ & $-0.56 \pm 0.01$ & $-0.56 \pm 0.01$ & $-0.56 \pm 0.01$ & $-0.56 \pm 0.01$ & $-0.55 \pm 0.01$ \\
\bottomrule
\end{tabular}
\end{scriptsize}
\end{table}

\newpage
\paragraph{Representation Quality.}
We next assess robustness to the quality and choice of the multimodal representations. First, we degrade representation quality by randomly masking a fraction of the embedding dimensions. Second, we replace the original ModernBERT embeddings with Gemma embeddings~\citep{vera2025embeddinggemma}. Tables~\ref{tab:synsum_masking_main} and~\ref{tab:synsum_gemma_main} show that (i) \textsc{AACE} degrades gradually under representation masking, and (ii) the relative ordering of methods remains largely unchanged with Gemma embeddings. This suggests that the benefits of \textsc{AACE} are not specific to the original encoder, but depend on the representations retaining relevant confounding signal.

\begin{table}[!h]
\centering
\caption{\textbf{\textsc{SynSum} (representation masking).} AUTOC of \textsc{AACE} as a function of the fraction of embedding dimensions randomly masked. See Table~\ref{tab:synsum_annotation_corruption_main} for details.}
\label{tab:synsum_masking_main}
\renewcommand{\arraystretch}{1.1}
\begin{scriptsize}
\begin{tabular}{l|ccccc}
\toprule
\textbf{Masking (\%)} & 0 & 75 & 90 & 95 & 100 \\
\midrule
\textbf{AUTOC} & $-0.56 \pm 0.01$ & $-0.54 \pm 0.01$ & $-0.51 \pm 0.01$ & $-0.46 \pm 0.00$ & $-0.32 \pm 0.01$ \\
\bottomrule
\end{tabular}
\end{scriptsize}
\end{table}

\begin{table}[!h]
\centering
\caption{\textbf{\textsc{SynSum} (alternative encoder).} AUTOC of all methods using ModernBERT and Gemma embeddings. See Table~\ref{tab:synsum_annotation_corruption_main} for details.}
\label{tab:synsum_gemma_main}
\renewcommand{\arraystretch}{1.1}
\begin{scriptsize}
\begin{tabular}{l|ccccc}
\toprule
\textbf{Encoder} & Risk & DR-Emb & DragonNet & TARNet & \textsc{AACE} \\
\midrule
ModernBERT & $-0.22 \pm 0.01$ & $-0.51 \pm 0.01$ & $-0.40 \pm 0.02$ & $-0.51 \pm 0.02$ & $\mathbf{-0.56 \pm 0.01}$ \\
Gemma & $-0.25 \pm 0.01$ & $-0.57 \pm 0.01$ & $-0.48 \pm 0.00$ & $-0.54 \pm 0.03$ & $\mathbf{-0.60 \pm 0.00}$ \\
\bottomrule
\end{tabular}
\end{scriptsize}
\end{table}

\subsection{Results on MIMIC-Real}

Table~\ref{tab:real_policy_main} reports agreement between the treatment rankings induced by \textsc{AACE}, TARNet, and the risk-based model on the \textsc{MIMIC-Real} dataset. As discussed above, these results do \textbf{not} allow us to determine which policy is best, since ground-truth treatment effects are unknown. Instead, they show that \textit{the different modeling strategies induce materially different treatment assignments in practice}. \textsc{AACE} is more strongly aligned with the representation-based causal baseline than with the risk-based model, indicating that causal and risk-based strategies prioritize substantially different patient groups in practice. Additional comparisons, including agreement with observed treatment assignments and the full pairwise analysis, are reported in Appendix~\ref{app:results}.

\begin{table}[!t]
\vspace{-0.4cm}
\centering
\caption{\textbf{\textsc{MIMIC-Real} (policy overlap).} Policy overlap and ranking agreement on real-world data (mean~$\pm$~std.~dev. over five random seeds). Jaccard measures agreement between top-$k$ treatment assignments, while Spearman $\rho$ captures global rank correlation.}
\label{tab:real_policy_main}
\renewcommand{\arraystretch}{1.2}
\begin{scriptsize}
\begin{tabular}{lcc}
\toprule
\textbf{Metric} & \textbf{AACE vs TARNet} & \textbf{AACE vs Risk}  \\
\midrule
\multicolumn{3}{c}{Jaccard overlap} \\
\midrule
$k = 5\%$  & $0.088 \pm 0.106$ & $0.037 \pm 0.027$ \\
$k = 10\%$ & $0.155 \pm 0.115$ & $0.076 \pm 0.048$ \\
$k = 20\%$ & $0.245 \pm 0.088$ & $0.141 \pm 0.052$ \\
$k = 50\%$ & $0.573 \pm 0.075$ & $0.362 \pm 0.051$ \\
\midrule
\multicolumn{3}{c}{Spearman correlation} \\
\midrule
$\rho$ & $0.551 \pm 0.202$ & $0.067 \pm 0.170$ \\
\bottomrule
\end{tabular}
\end{scriptsize}
\end{table}

\section{Discussion} 

\paragraph{Empirical Findings.}
Across synthetic and semi-synthetic benchmarks, \textsc{AACE} outperforms both risk-based and representation-based causal baselines, and remains robust to moderately imperfect annotations and representations. On real-world EHR data, the different methods induce substantially different treatment assignments, suggesting that the choice of modeling strategy can directly affect which patients are prioritized for treatment. Together, these results show that targeted training-time supervision can substantially improve treatment policy learning in multimodal settings where representations alone are insufficient for confounding adjustment.

\paragraph{Technical Implications.}
We study an important yet understudied setting in causal inference: multimodal data, where learned representations may not be sufficient for confounding adjustment. In that case, treatment effects are \textit{not} identified from the observed data. Applying causal estimators directly to the representations may still perform well empirically, but the resulting treatment effect estimates remain \textit{biased}, and the magnitude of this bias is generally unknown. In this setting, confounding must be addressed using additional training-time covariates that are sufficient for confounding adjustment. We argue that expert annotations offer a practical way to obtain such covariates.

\paragraph{Clinical Relevance.}
Electronic health records are a natural choice for learning data-driven treatment policies, because they are widely available and contain rich patient-specific information. However, learning \textit{effective} and \textit{reliable} treatment policies requires proper confounding adjustment, which is non-trivial in multimodal EHRs where relevant confounders may be distributed across tabular data and text. With \textsc{AACE}, we argue that expert knowledge remains important in this setting. By using expert-provided annotations during training, \textsc{AACE} offers a practical way to support confounding adjustment while still operating on standard multimodal EHR inputs at inference. This makes the approach compatible with standard clinical workflows, since the expert input is required only during model development, while routine EHR data are used for deployment. As with any clinical decision-support model, applying \textsc{AACE} in a new setting requires careful evaluation of the learned policy in the target population and clinical context.

\paragraph{Limitations and Future Work.}
Our study has several limitations. First, as noted in Section~\ref{sec:problem} and illustrated in Appendix~\ref{app:annotations}, we do not propose a universal annotation protocol for multimodal EHRs. This is deliberate, since the required training-time covariates for confounding adjustment depend on the healthcare application studied. Which covariates require annotation in practice should be guided by domain expertise. Second, as is standard in causal inference, our quantitative evaluation is limited to synthetic and semi-synthetic benchmarks where ground-truth treatment effects are available, while the analysis on real-world data remains qualitative. A useful direction for future work is therefore to evaluate \textsc{AACE} on RCT data, where policy performance can be measured from observed outcomes. Third, our experiments rely on simulated annotations or proxy training-time covariates rather than expert-provided annotations. An important next step is therefore to evaluate the method with real annotations, as well as more scalable forms of supervision such as partial annotation or LLM-assisted labeling with expert verification.

\section{Conclusion}
In this work, we study treatment policy learning from multimodal electronic health records (EHRs), where confounding information is distributed across tabular data and clinical text. We proposed \textsc{AACE}, an annotation-assisted method that uses expert-provided annotations during training to support confounding adjustment, while relying only on multimodal representations at inference time. Across synthetic and semi-synthetic benchmarks, \textsc{AACE} outperforms both risk-based and representation-based causal baselines, while remaining robust to moderately imperfect annotations and representations. On real-world data, we find that the choice of modeling strategy substantially affects the resulting treatment assignments, highlighting that method choice matters in practice. Together, these results suggest that incorporating expert knowledge during training is a promising direction for improving treatment policy learning from multimodal healthcare data.

\acks{This work received funding from the Research Foundation Flanders (FWO Vlaanderen) with grant
number 11Q2C24N and from the Flemish government under the ``Onderzoeksprogramma Artificiele Intelligentie (AI)
Vlaanderen" program.}

\clearpage
\bibliography{biblio}

\clearpage
\appendix

\section{Annotation Procedure}
\label{app:annotations}

As discussed in Section~\ref{sec:problem}, when clinically relevant confounders are expressed in text but not reliably preserved in the learned representation $\phi$, treatment effects are not identified from observational data $(\phi, T, Y)$ alone. Treatment effect estimation in this multimodal setting therefore requires training-time covariates that are sufficient for confounding adjustment. With AACE, we argue that expert-provided annotations offer a pragmatic way to obtain such covariates. Because the specific variables that need to be annotated depend on the clinical application and require domain expertise, we deliberately do not prescribe a fixed annotation procedure. Instead, we provide below an illustrative example of how such annotations may be obtained in one of our datasets.

\begin{figure}[h]
    \centering
    \includegraphics[width=\columnwidth]{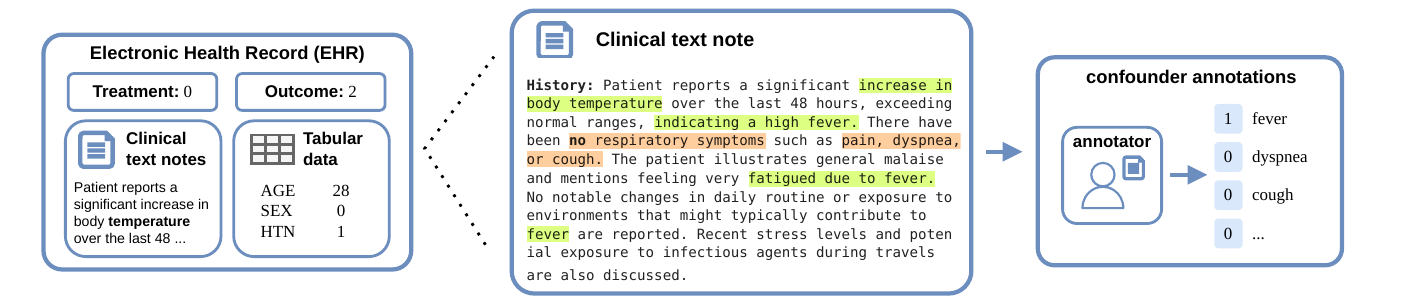}
    \vspace{-0.7cm}
    \caption{\textbf{SynSum (example annotation procedure).}
    Illustration of how clinically relevant confounding information may be expressed implicitly in a clinical note and annotated for use in AACE. In this example, highlighted spans provide evidence for symptom variables such as \textit{fever}, \textit{dyspnea}, and \textit{cough}, which are used as training-time covariates for confounding adjustment.}
    \label{fig:annotations}
    \vspace{-0.5cm}
\end{figure}

\paragraph{Problem Setting.}
\textsc{SynSum} is a dataset that simulates primary-care encounters for patients with respiratory diseases (discussed in detail in Appendix~\ref{app:datasets}). Each simulated record contains both tabular variables and a free-text clinical note: background information, such as demographics, is recorded in structured form, while symptom variables such as \textit{fever}, \textit{dyspnea}, and \textit{cough} are expressed in the note. In this setting, the treatment decision under study is antibiotics prescription, which is influenced by the symptoms that also affect patient outcomes, and therefore requires adjustment for them during effect estimation.

\paragraph{Example Annotation.}
Figure~\ref{fig:annotations} illustrates one plausible \textit{retrospective} annotation workflow for AACE, in which a domain expert identifies clinical factors from the note that may influence both antibiotic prescription and patient outcomes, and records them in structured form. In \textsc{SynSum}, this corresponds to the symptom variables. The highlighted spans in Figure~\ref{fig:annotations} indicate the textual evidence supporting these variables, while the binary indicators illustrate how they may be represented as additional training-time covariates. In practice, the same variables could also be recorded directly as part of routine clinical documentation during \textit{a one-time annotation effort}. In that case, the same domain expertise used to assess the patient and make the treatment decision could also be used to record a small set of structured covariates that are relevant for confounding adjustment. These covariates are then used, together with the tabular EHR variables, for confounding adjustment during training. At inference time, the annotations are not required and the policy depends only on the representation $\phi$.

\clearpage
\section{Proof}
\label{app:ordering}

In this appendix, we provide the proof of Proposition~\ref{prop:topk}.\\

\begin{proof}
Assume that the true treatment effects have no ties. Let $x_{(1)}, \dots, x_{(N)}$ denote the individuals ordered by decreasing treatment effects, so that
\begin{equation}
\tau^x(x_{(1)}) > \tau^x(x_{(2)}) > \cdots > \tau^x(x_{(N)}).
\end{equation}
For a fixed treatment budget $k < N$, define the treatment effect margin at the decision threshold as
\begin{equation}
\gamma_k = \tau^x(x_{(k)}) - \tau^x(x_{(k+1)}),
\end{equation}
and let
\begin{equation}
\delta_i = \bigl| \tau^x(x_i) - \tau^\phi(\phi_i) \bigr|
\end{equation}
denote the coarsening bias for individual $i$.\\

Let $S_k = \{x_{(1)}, \dots, x_{(k)}\}$ denote the true top-$k$ according to $\tau^x(x)$. To prove the proposition, it suffices to show that every individual in $S_k$ remains ranked above every individual outside $S_k$ when ranking by the coarsened effects $\tau^\phi(\phi)$. Now consider any $x_i \in S_k$ and any $x_j \notin S_k$. By construction we have,
\begin{equation}
\tau^x(x_i) \ge \tau^x(x_{(k)})
\qquad \text{and} \qquad
\tau^x(x_j) \le \tau^x(x_{(k+1)}).
\end{equation}
Therefore,
\begin{align}
\tau^x(x_i) - \tau^x(x_j)
&\ge
\tau^x(x_{(k)}) - \tau^x(x_{(k+1)})\\
&\ge
\gamma_k.
\end{align}
By definition of the coarsening bias,
\begin{equation}
\tau^\phi(\phi_i) \ge \tau^x(x_i) - \delta_i
\qquad \text{and} \qquad
\tau^\phi(\phi_j) \le \tau^x(x_j) + \delta_j.
\end{equation}
Subtracting yields
\begin{align}
\tau^\phi(\phi_i) - \tau^\phi(\phi_j)
&\ge \big(\tau^x(x_i) - \delta_i \big) - \bigl(\tau^x(x_j) + \delta_j\bigr) \\
&= \bigl(\tau^x(x_i) - \tau^x(x_j)\bigr) - (\delta_i + \delta_j) \\
&\ge \gamma_k - 2 \max_m \delta_m.
\end{align}
By the assumption of the proposition,
\begin{equation}
2 \max_m \delta_m < \gamma_k,
\end{equation}
and therefore
\begin{equation}
\tau^\phi(\phi_i) - \tau^\phi(\phi_j) > 0.
\end{equation}
Hence,
\begin{equation}
\tau^\phi(\phi_i) > \tau^\phi(\phi_j).
\end{equation}

Since this holds for every $x_i \in S_k$ and every $x_j \notin S_k$, every true top-$k$ individual remains ranked above every individual outside the top-$k$ under the coarsened effects. Therefore, ranking by $\tau^\phi(\phi)$ yields exactly the same top-$k$ treatment set as ranking by $\tau^x(x)$.
\end{proof}

\clearpage
\section{Datasets}
\label{app:datasets}
This appendix provides a detailed description of the datasets used in our experiments: the fully synthetic \textsc{SynSum} dataset \citep{SynSUM}, the semi-synthetic \textsc{MIMIC-Syn} dataset \citep{johnson2016mimic, chen2024proximal}, and the real-world \textsc{MIMIC-Real} dataset \citep{johnson2016mimic}. For the synthetic and semi-synthetic datasets, we describe the data-generating process, including the construction of treatments and outcomes. The \textsc{MIMIC-Real} dataset is included \textbf{only} for qualitative comparison of the treatment assignment strategies, since ground-truth treatment effects are unknown.

\subsection{SynSum}
\label{app:simsum}
The fully synthetic \textsc{SynSum} dataset~\citep{SynSUM}\footnote{\url{https://github.com/prabaey/SimSum}} consists of 10{,}000 medical patient records simulating primary care encounters for patients with respiratory diseases. Each record combines tabular variables with an associated free-text clinical note, mimicking the multimodal structure of real electronic health records (EHRs). The dataset is entirely specified and reproducible: tabular variables are sampled from a Bayesian network defined by a domain expert, while clinical notes are generated using GPT-4. \textsc{SynSum} thus provides realistic, simulated EHR data with ground-truth counterfactuals, enabling controlled evaluation of the considered treatment assignment strategies.

\paragraph{Tabular Variables.}
In \textsc{SynSum}, the structured variables are sampled from a Bayesian network that defines the underlying clinical dependencies in the simulated population. The variables include diagnoses (pneumonia and common cold), symptoms (dyspnea, cough, pain, fever, and nasal congestion), underlying respiratory conditions (asthma, smoking, chronic obstructive pulmonary disease, and hay fever), and non-clinical factors (season, prescription policy, and self-employment). Each variable is binary and represented by an indicator $x_{\text{var}} \in \{0,1\}$. Full details of the Bayesian network and its parametrization are given in~\citet{SynSUM}. The structured variables play different roles in the benchmark. A subset of them (self-employment status, smoking, winter season, asthma, chronic obstructive pulmonary disease, and hay fever) appears as tabular variables in the simulated EHRs, representing background and chronic-condition information typically recorded in structured form. The symptom variables (dyspnea, cough, pain, fever, and nasal congestion) are used to generate the accompanying clinical text, where they are expressed implicitly rather than as structured fields. Finally, a small number of variables (the diagnoses and the policy indicator) are used only to induce realistic dependencies in the data-generating process and are not included directly in the final EHR.

\paragraph{Treatment and Outcome.}
Based on the structured variables, we simulate treatment assignment and potential outcomes. Relative to the original \textsc{SynSum} data, we define a new treatment and outcome generating process tailored to treatment policy learning, with stronger confounding and weaker alignment between baseline risk and treatment benefit. The treatment variable $T \in \{0,1\}$ indicates whether antibiotics are prescribed during the simulated consultation. It is sampled from a logistic regression model
\begin{equation}
P(T=1 \mid X=x) = \sigma\bigl(\ell(x)\bigr),
\end{equation}
where $\sigma(\cdot)$ denotes the sigmoid function and the logit is given by
\begin{align}
\ell(x) =\;&
-2.2
+ 2.0\,x_{\text{pol}}
+ 1.5\,x_{\text{dysp}}
+ 1.1\,x_{\text{cough}}
+ 1.3\,x_{\text{f\_high}}
+ 0.6\,x_{\text{pain}} \nonumber \\ 
&+ 0.9\,x_{\text{pol}}\,x_{\text{dysp}}
+ 0.7\,x_{\text{pol}}\,x_{\text{f\_high}}
- 0.5\,x_{\text{nasal}}\,x_{\text{f\_low}}.
\end{align}
This specification reflects realistic clinical behaviour, where treatment is more likely for patients with more severe symptoms, especially under permissive prescription policies. We then define the untreated potential outcome through a baseline response surface $\mu_0(x)$,
\begin{equation}
\begin{aligned}
\mu_0(x) =\;&
1.8\,x_{\text{self}}
+ 1.6\,x_{\text{dysp}}
+ 1.2\,x_{\text{cough}}
+ 0.5\,x_{\text{pain}}
+ 0.4\,x_{\text{nasal}} \\
&+ 0.3\,x_{\text{f\_low}}
+ 1.5\,x_{\text{f\_high}}
+ 1.1\,x_{\text{dysp}}\,x_{\text{f\_high}} \\
&+ 0.9\,x_{\text{self}}\,x_{\text{cough}}
- 0.6\,x_{\text{nasal}}\,x_{\text{f\_low}} \\
&+ 0.7\,x_{\text{pain}}\,x_{\text{self}}
+ 0.8\,x_{\text{cough}}\,x_{\text{f\_high}}
+ 0.7\,x_{\text{dysp}}\,x_{\text{self}}.
\end{aligned}
\end{equation}
The conditional average treatment effect (CATE) is specified as
\begin{equation}
\tau^x(x) =
-1
+ 1.1\,x_{\text{pain}}
+ 1.0\,x_{\text{nasal}}
+ 0.7\,x_{\text{f\_low}}
- 1.0\,x_{\text{dysp}}
- 0.7\,x_{\text{cough}}
- 1.1\,x_{\text{f\_high}},
\end{equation}
so that treatment is more beneficial for patients with more severe lower-respiratory symptoms, and less beneficial for those with milder or upper-airway symptoms. The treated response surface is then defined as
\begin{equation}
\mu_1(x) = \mu_0(x) + \tau^x(x).
\end{equation}
Finally, the potential outcomes are generated by adding Gaussian noise,
\begin{equation}
Y(0) = \mu_0(x) + \varepsilon_0, \qquad
Y(1) = \mu_1(x) + \varepsilon_1,
\end{equation}
where $\varepsilon_0, \varepsilon_1 \sim \mathcal{N}(0,\sigma_y^2)$, and the observed outcome is
\begin{equation}
Y = T\,Y(1) + (1-T)\,Y(0).
\end{equation}
This construction ensures that symptom variables act as confounders by influencing both treatment assignment and outcome, while treatment benefit is heterogeneous and only partially aligned with baseline risk.

\paragraph{Text Generation.}
Each record’s clinical note is generated with GPT-4 so that the patient’s symptom profile is reflected in natural language. The prompts are designed to encourage clinical realism, coherence, and stylistic variation, resembling documentation written at the time of consultation. In the final EHR, these symptoms are not included as structured variables, but are expressed implicitly in the free-text note. Each simulated EHR therefore consists of tabular background variables (e.g., smoking status, chronic conditions, and season) together with a clinical note describing the patient’s symptoms.

\paragraph{Feature Representations.}
Each simulated EHR in \textsc{SynSum} consists of tabular background variables and a clinical note. We encode the text using the pretrained language model ModernBERT~\citep{modernbert}, mean-pool the token embeddings, and concatenate the resulting text representation with the tabular variables to obtain the multimodal representation~$\phi$. This representation combines the structured background information with confounding signal expressed implicitly in the text.

\paragraph{Confounding Annotations.}
\textsc{AACE} assumes that domain experts can provide annotations of confounding information expressed in the clinical text. In \textsc{SynSum}, this process is simulated using the ground-truth symptom variables (e.g., fever, cough, and dyspnea), which influence both treatment assignment and outcome. This yields a controlled setting in which confounding adjustment is possible during training through the annotations, while inference relies only on the multimodal representations. In addition, we use this benchmark later to study how \textsc{AACE} degrades as annotation quality decreases.

\subsection{MIMIC-Syn}
\label{app:mimicsyn}

The semi-synthetic \textsc{MIMIC-Syn} dataset is derived from the publicly available \textsc{MIMIC-III} database~\citep{johnson2016mimic}\footnote{\url{https://physionet.org/content/mimiciii/1.4/}}, which contains de-identified electronic health records from over 35{,}000 intensive care unit (ICU) patients. The dataset includes both structured variables and free-text clinical notes. Following~\citet{chen2024proximal}, we retain the real tabular and textual features while generating synthetic treatment and outcome variables. This design preserves the realism of real-world EHR data while enabling controlled evaluation under a known causal ground truth.

\paragraph{Data Preprocessing.}
We follow the preprocessing pipeline of~\citet{chen2024proximal} to construct patient-level records from \textsc{MIMIC-III}. Starting from all clinical notes, we remove entries with missing hospital admission IDs and exclude discharge summaries. For each patient, identified by a unique subject ID, we retain only the earliest hospital admission and concatenate all associated clinical notes into a single text sequence. We additionally extract the demographic variables age and sex, which serve as the tabular covariates in the EHR. Patients younger than 18 or older than 100 years are excluded. Each resulting record therefore consists of a single aggregated clinical note and a small set of demographic variables, forming a multimodal EHR with real-world patient data. Other structured variables from \textsc{MIMIC-III} are used later in the data-generating process to construct treatment and outcome variables, but are not included directly in the final EHR.

\paragraph{Treatment and Outcome.}
To introduce heterogeneous treatment effects, we define a semi-synthetic treatment and outcome generating process on top of the real \textsc{MIMIC-III} covariates. Treatment assignment and potential outcomes are generated from four cardiovascular diagnoses (hypertension, coronary atherosclerosis, atrial fibrillation, and congestive heart failure) together with age and sex. The diagnosis indicators ($x_{\text{var}} \in \{0,1\}$) are extracted from the ICD-9 codes in the \textsc{MIMIC-III} tables, but are not included in the constructed EHRs. Prior analyses by~\citet{chen2024proximal} showed that these diagnoses are among the most predictable conditions from clinical text, indicating that they are implicitly expressed in the notes. Using these variables to generate treatments and outcomes therefore induces implicit multimodal confounding, analogous to the \textsc{SynSum} setting. The treatment variable $T \in \{0,1\}$ indicates whether the hypothetical intervention is assigned. It is sampled from a logistic regression model
\begin{equation}
P(T=1 \mid X=x) = \sigma\bigl(\ell(x)\bigr),
\end{equation}
where $\sigma(\cdot)$ denotes the sigmoid function and the logit is given by
\begin{equation}
\begin{aligned}
\ell(x) =\;&
-0.8
+ 0.08\,x_{\text{age,c}}
+ 0.65\,x_{\text{sex}}
+ 0.50\,x_{\text{hyp}}
+ 0.60\,x_{\text{cor}} \\
&+ 0.40\,x_{\text{afib}}
+ 0.70\,x_{\text{chf}}
+ 0.25\,x_{\text{sex}}\,\mathbb{I}(x_{\text{age}}>75) \\
&+ 0.25\,x_{\text{cor}}\,x_{\text{chf}},
\end{aligned}
\end{equation}
with $x_{\text{age,c}} = x_{\text{age}} - 65$. This specification reflects confounding by indication, where older patients and patients with greater cardiovascular burden are more likely to receive treatment, while also being at higher baseline risk. We define the untreated potential outcome through a baseline response surface $\mu_0(x)$,
\begin{equation}
\begin{aligned}
\mu_0(x) =\;&
5.5
+ 0.10\,x_{\text{age,c}}
+ 0.60\,x_{\text{sex}}
+ 0.70\,x_{\text{hyp}}
+ 1.00\,x_{\text{cor}}
+ 0.85\,x_{\text{afib}}
+ 1.20\,x_{\text{chf}} \\
&+ 0.020\,\max(x_{\text{age}}-75,0)^2
+ 0.50\,x_{\text{cor}}\,x_{\text{chf}}
+ 0.35\,x_{\text{afib}}\,x_{\text{chf}} \\
&+ 0.25\,x_{\text{hyp}}\,x_{\text{cor}}
+ 0.20\,x_{\text{sex}}\,x_{\text{cor}}
+ 0.015\,x_{\text{age,c}}\,x_{\text{afib}}.
\end{aligned}
\end{equation}
The conditional average treatment effect (CATE) is specified as
\begin{equation}
\begin{aligned}
\tau^x(x) =\;&
-1.0
- 0.35\,x_{\text{hyp}}
- 0.50\,x_{\text{cor}}
+ 0.20\,x_{\text{afib}}
+ 0.45\,x_{\text{chf}} \\
&+ 0.015\,\max(x_{\text{age}}-75,0)
+ 0.05\,x_{\text{sex}}
+ 0.20\,x_{\text{afib}}\,x_{\text{chf}}
- 0.35\,x_{\text{hyp}}\,x_{\text{cor}}.
\end{aligned}
\end{equation}
Negative values of $\tau^x(x)$ indicate treatment benefit, since lower outcomes are better. This specification makes treatment beneficial on average, while allowing the benefit to vary substantially across subgroups. The treated response surface is then defined as
\begin{equation}
\mu_1(x) = \mu_0(x) + \tau^x(x).
\end{equation}
Finally, the potential outcomes are generated by adding Gaussian noise,
\begin{equation}
Y(0) = \mu_0(x) + \varepsilon_0, \qquad
Y(1) = \mu_1(x) + \varepsilon_1,
\end{equation}
where $\varepsilon_0, \varepsilon_1 \sim \mathcal{N}(0,\sigma_y^2)$, and the observed outcome is
\begin{equation}
Y = T\,Y(1) + (1-T)\,Y(0).
\end{equation}
This setup produces a semi-synthetic benchmark in which treatment assignment is confounded, treatment benefit is heterogeneous, and baseline risk is not sufficient to determine which patients benefit most from treatment.

\paragraph{Feature Representations.}
Each record in \textsc{MIMIC-Syn} consists of a clinical note and tabular demographic variables (age and sex). We encode the clinical text using the pretrained language model ModernBERT~\citep{modernbert}, mean-pool the token embeddings, and concatenate the resulting text representation with the tabular covariates to obtain the multimodal representation~$\phi$. This representation combines the demographic information with implicit confounding signal related to the latent cardiovascular diagnoses. Consequently, $\phi$ provides a coarsened view of the underlying factors influencing treatment and outcome, mirroring the partial observability typical of real-world EHR data.

\paragraph{Confounding Annotations.}
\textsc{AACE} assumes that domain experts can provide annotations of confounding information expressed in the clinical text. In \textsc{MIMIC-Syn}, this process is simulated using the ground-truth cardiovascular diagnoses (hypertension, coronary atherosclerosis, atrial fibrillation, and congestive heart failure), which influence both treatment assignment and outcome. This yields a controlled semi-synthetic setting in which confounding adjustment is possible during training through the annotations, while inference relies only on the multimodal representations~$\phi$.

\subsection{MIMIC-Real}
\label{app:mimicreal}
We construct a real-world observational dataset from the \textsc{MIMIC-III} database~\citep{johnson2016mimic} to examine whether annotation-assisted, risk-based, and representation-based causal treatment strategies diverge in practice. In this setting, both the tabular variables and the clinical text notes of the electronic health records (EHRs) are derived directly from real-world patient data. The treatment (\textit{antibiotic administration}) and outcome (\textit{in-hospital mortality}) are likewise \textbf{observed} rather than generated. This dataset complements the synthetic and semi-synthetic analyses by extending the comparison of risk-based and causal policies to fully real-world clinical data. Because ground-truth counterfactuals are unknown, \textbf{we cannot determine which policy is best} and restrict the analysis to a \textbf{qualitative comparison} of the treatment assignments induced by the different methods.

\paragraph{Data Preprocessing.}
We construct a patient-level dataset of EHRs from all intensive care unit (ICU) stays in \textsc{MIMIC-III}. Each record combines tabular variables and clinical notes describing the patient’s state prior to treatment. We define a reference time $t_0$ three hours after ICU admission and a follow-up window $\Delta t$ of six hours. All covariates are derived from data recorded up to $t_0$; treatment is defined within the interval $(t_0,\, t_0 + \Delta t)$; and the outcome is measured after $t_0 + \Delta t$. ICU stays that end before $t_0 + \Delta t$ are excluded. As tabular variables, we retain the demographic attributes age and sex, consistent with \textsc{MIMIC-Syn}. We exclude individuals younger than 18 years and retain only the first ICU stay of each patient. All clinical notes recorded before $t_0$ are extracted and concatenated into a single text sequence representing the documentation available at the time of the treatment decision. Each resulting EHR therefore consists of the tabular demographic variables age and sex together with an aggregated clinical note summarizing the patient’s condition prior to treatment.

\paragraph{Treatment and Outcome.}
The treatment variable $T \in \{0,1\}$ indicates whether at least one antibiotic prescription was administered during the interval $(t_0,\, t_0 + \Delta t)$. Antibiotic prescriptions are identified using a list of antibiotic drug names. The outcome variable $Y \in \{0,1\}$ denotes in-hospital mortality occurring after $t_0 + \Delta t$, reflecting post-treatment mortality between the end of the treatment window and hospital discharge. In the resulting dataset (8{,}685 ICU stays), 18.5\% of patients received antibiotics within the treatment window ($T{=}1$), and 12.0\% died after the treatment window during the same hospitalization ($Y{=}1$). Mortality was 16.4\% among treated patients and 11.0\% among untreated patients. This higher mortality among treated patients does not imply a harmful treatment effect, but instead reflects \textit{confounding}: antibiotics are typically administered to patients with suspected sepsis or severe infection, who are at higher risk of death. Treatment and outcome are therefore strongly associated through shared confounders such as abnormal vital signs.

\paragraph{Feature Representations.}
Each EHR in \textsc{MIMIC-Real} consists of the tabular demographic variables age and sex together with an aggregated clinical note describing the patient’s condition prior to treatment. We encode the clinical text using the pretrained language model ModernBERT~\citep{modernbert}, mean-pool the token embeddings, and concatenate the resulting text representation with the tabular variables to obtain the multimodal representation~$\phi$. This representation combines explicit demographic information with implicit clinical signals present in the real-world EHR data.

\paragraph{Additional Training-Time Covariates.}
For \textsc{AACE}, we use pre-treatment vital signs as additional training-time covariates for confounding adjustment. These measurements plausibly influence both antibiotic administration and mortality: clinicians are more likely to prescribe antibiotics to patients presenting with abnormal physiological measurements, such as fever, tachycardia, or hypotension, and these same measurements are also associated with higher mortality risk. We therefore extract the following vital signs prior to treatment (before $t_0$): temperature, heart rate, respiratory rate, oxygen saturation, systolic blood pressure, and mean arterial pressure.\\

\noindent
To assess whether this confounding information is implicitly expressed in the clinical text, we evaluate how well abnormal vital-sign patterns can be predicted from the text representations. Vital signs are discretized based on standard clinical thresholds: fever (temperature $\ge 37.5^\circ$C), hypothermia (temperature $\le 35.5^\circ$C), tachycardia (heart rate $\ge 100$ beats/min), tachypnea (respiratory rate $\ge 22$ breaths/min), hypoxemia (oxygen saturation $< 92\%$), and hypotension (mean arterial pressure $< 65$ mmHg or systolic blood pressure $< 90$ mmHg). Each abnormality indicator is then predicted using only the ModernBERT text embeddings. The dataset is randomly divided into 80\% training, 10\% validation, and 10\% test splits. As shown in Table~\ref{tab:vitals_text_predictability}, these physiological states are predictable from the clinical text, with all AUROC values exceeding 0.5. This indicates that clinically relevant confounding information is at least partly encoded in the notes. Accordingly, we use the continuous vital-sign measurements as additional training-time covariates for confounding adjustment during training of \textsc{AACE}. At inference time, the method relies only on the multimodal representations~$\phi$. We note, however, that this real-world dataset may still contain additional \textbf{unobserved confounders} beyond the available vital-sign measurements.

\subsection{Summary}
In summary, the three datasets (\textsc{SynSum}, \textsc{MIMIC-Syn}, and \textsc{MIMIC-Real}) enable evaluation of annotation-assisted, risk-based, and representation-based causal treatment assignment strategies across settings ranging from fully synthetic simulations to real-world clinical data. In the following section, we present additional experimental results and analyses based on these datasets.

\begin{table*}[ht]
\centering
\renewcommand{\arraystretch}{1}
\begin{footnotesize}
\caption{\textbf{\textsc{MIMIC-Real} (confounder predictability from text).} Prediction of physiological confounders from pre-$t_0$ clinical notes (mean $\pm$ std. over five random seeds). ``Prev.'' denotes the positive-class prevalence in the test set. Each classifier's decision threshold was selected to maximize F1 score on the validation set.}
\label{tab:vitals_text_predictability}
\begin{tabular}{lccccc}
\toprule
\textbf{Confounder} & \textbf{Prev.} & \textbf{F1} & \textbf{Acc.} & \textbf{AUPRC} & \textbf{AUROC} \\
\midrule
Fever      & 0.14 & $0.337 \pm 0.007$ & $0.607 \pm 0.026$ & $0.253 \pm 0.003$ & $0.680 \pm 0.003$ \\
Hypothermia             & 0.04 & $0.026 \pm 0.041$ & $0.951 \pm 0.013$ & $0.097 \pm 0.026$ & $0.649 \pm 0.024$ \\
Tachycardia             & 0.22 & $0.435 \pm 0.006$ & $0.617 \pm 0.021$ & $0.381 \pm 0.002$ & $0.679 \pm 0.003$ \\
Tachypnea               & 0.26 & $0.467 \pm 0.006$ & $0.599 \pm 0.025$ & $0.408 \pm 0.004$ & $0.678 \pm 0.005$ \\
Hypoxemia               & 0.03 & $0.111 \pm 0.012$ & $0.746 \pm 0.093$ & $0.118 \pm 0.006$ & $0.629 \pm 0.010$ \\
Hypotension (MAP)       & 0.17 & $0.343 \pm 0.006$ & $0.539 \pm 0.011$ & $0.278 \pm 0.011$ & $0.636 \pm 0.007$ \\
Hypotension (SBP)       & 0.06 & $0.200 \pm 0.013$ & $0.790 \pm 0.017$ & $0.130 \pm 0.004$ & $0.714 \pm 0.006$ \\
\bottomrule
\end{tabular}
\end{footnotesize}
\end{table*}

\newpage
\section{Extended Experimental Results}
\label{app:results}
This appendix provides extended experimental results that complement the main text. We report extended evaluations on \textsc{SynSum} and \textsc{MIMIC-Syn}, including robustness analyses and ablations, as well as additional analysis of treatment assignments on \textsc{MIMIC-Real}.

\subsection{SynSum}
\label{app:synsum_results}

\paragraph{Recoverability of Confounders From Text.}
A key assumption underlying representation-based causal methods is that the learned representations are sufficient for confounding adjustment. To assess this in \textsc{SynSum}, we evaluate how well the symptom variables used in the data-generating process can be recovered from the text representations. Specifically, we train classifiers to predict each symptom variable from the clinical notes. As shown in Table~\ref{tab:simsum_confounders}, the confounders are predictable from text, although imperfectly, confirming that the notes contain substantial confounding signal.

\begin{table*}[ht]
\centering
\begin{footnotesize}
\caption{\textbf{\textsc{SynSum} (confounder predictability from text).} Prediction of symptom confounders from clinical notes (mean $\pm$ std. over five random seeds). ``Prev.'' denotes the positive-class prevalence in the test set. Each classifier's decision threshold was selected to maximize F1 score on the validation set.}
\renewcommand{\arraystretch}{1}
\label{tab:simsum_confounders}
\begin{tabular}{lcccccc}
\toprule
\textbf{Confounder} & \textbf{Prev.} & \textbf{F1} & \textbf{Acc.} & \textbf{AUPRC} & \textbf{AUROC} \\
\midrule
Cough         & 0.36 & $0.971 \pm 0.001$ & $0.979 \pm 0.001$ & $0.989 \pm 0.000$ & $0.991 \pm 0.000$ \\
Dyspnea       & 0.20 & $0.949 \pm 0.010$ & $0.980 \pm 0.004$ & $0.980 \pm 0.001$ & $0.990 \pm 0.001$ \\
Fever (high)  & 0.07 & $0.995 \pm 0.007$ & $0.999 \pm 0.001$ & $1.000 \pm 0.000$ & $1.000 \pm 0.000$ \\
Fever (low)   & 0.17 & $0.841 \pm 0.006$ & $0.951 \pm 0.002$ & $0.879 \pm 0.001$ & $0.935 \pm 0.000$ \\
Fever (none)  & 0.77 & $0.968 \pm 0.001$ & $0.950 \pm 0.002$ & $0.983 \pm 0.000$ & $0.956 \pm 0.001$ \\
Nasal symptoms & 0.25 & $0.974 \pm 0.001$ & $0.987 \pm 0.000$ & $0.984 \pm 0.000$ & $0.990 \pm 0.000$ \\
Pain          & 0.15 & $0.796 \pm 0.014$ & $0.950 \pm 0.003$ & $0.851 \pm 0.006$ & $0.930 \pm 0.004$ \\
\bottomrule
\end{tabular}
\end{footnotesize}
\end{table*}

\newpage
\paragraph{Policy Performance Across Training Sizes.}
In addition to the main results, we report policy evaluation metrics across a wider range of training sizes (Table~\ref{tab:synsum_policyapp}). As the number of training samples increases, all methods improve, but \textsc{AACE} consistently achieves better policy performance than both risk-based and representation-based causal baselines.

\begin{table*}[h]
\centering
\caption{\textbf{\textsc{SynSum} (main results).} Policy metrics across training sizes, evaluated on a fixed test set (mean~$\pm$~std.~dev. over five random seeds). Lower values are better. PEHE measures treatment effect estimation error and is not defined for the risk-based model. The best model within each training size is shown in \textbf{bold}. The \textit{oracle} row reports the metrics of a policy that ranks by the true treatment effect.}
\label{tab:synsum_policyapp}
\renewcommand{\arraystretch}{1}
\begin{footnotesize}
\begin{tabular}{lccccc}
\toprule
\textbf{Training size} & \textbf{Model} & \textbf{AUTOC $(\downarrow)$} & \textbf{$\boldsymbol{\overline{V}}$ $(\downarrow)$} & \textbf{$\boldsymbol{V_{10\%}}$ $(\downarrow)$} & \textbf{PEHE $(\downarrow)$} \\
\midrule
\multirow{5}{*}{1{,}000}
 & Risk & $-0.25 \pm 0.02$ & $0.77 \pm 0.01$ & $1.17 \pm 0.01$ & --- \\
 & DR-Emb & $-0.42 \pm 0.04$ & $0.69 \pm 0.01$ & $1.14 \pm 0.01$ & $0.71 \pm 0.02$ \\
 & DragonNet & $-0.09 \pm 0.16$ & $0.79 \pm 0.04$ & $1.21 \pm 0.04$ & $0.96 \pm 0.02$ \\
 & TARNet & $-0.43 \pm 0.03$ & $0.69 \pm 0.01$ & $1.14 \pm 0.01$ & $0.70 \pm 0.03$ \\
 & AACE (ours) & $\mathbf{-0.49 \pm 0.02}$ & $\mathbf{0.68 \pm 0.01}$ & $\mathbf{1.12 \pm 0.01}$ & $\mathbf{0.62 \pm 0.02}$ \\
\midrule
\multirow{5}{*}{2{,}000}
 & Risk & $-0.23 \pm 0.03$ & $0.77 \pm 0.01$ & $1.18 \pm 0.01$ & --- \\
 & DR-Emb & $-0.46 \pm 0.02$ & $0.68 \pm 0.00$ & $1.13 \pm 0.01$ & $0.64 \pm 0.02$ \\
 & DragonNet & $-0.10 \pm 0.06$ & $0.77 \pm 0.02$ & $1.22 \pm 0.01$ & $0.83 \pm 0.04$ \\
 & TARNet & $-0.45 \pm 0.03$ & $0.68 \pm 0.01$ & $1.14 \pm 0.01$ & $0.67 \pm 0.03$ \\
 & AACE (ours) & $\mathbf{-0.53 \pm 0.01}$ & $\mathbf{0.66 \pm 0.00}$ & $\mathbf{1.12 \pm 0.00}$ & $\mathbf{0.57 \pm 0.01}$ \\
\midrule
\multirow{5}{*}{4{,}000}
 & Risk & $-0.22 \pm 0.01$ & $0.77 \pm 0.00$ & $1.18 \pm 0.00$ & --- \\
 & DR-Emb & $-0.51 \pm 0.01$ & $0.66 \pm 0.00$ & $1.12 \pm 0.00$ & $0.59 \pm 0.02$ \\
 & DragonNet & $-0.40 \pm 0.02$ & $0.70 \pm 0.00$ & $1.14 \pm 0.01$ & $0.71 \pm 0.02$ \\
 & TARNet & $-0.51 \pm 0.02$ & $0.66 \pm 0.00$ & $1.13 \pm 0.01$ & $0.58 \pm 0.02$ \\
 & AACE (ours) & $\mathbf{-0.56 \pm 0.01}$ & $\mathbf{0.65 \pm 0.00}$ & $\mathbf{1.11 \pm 0.01}$ & $\mathbf{0.53 \pm 0.02}$ \\
\midrule
\multirow{5}{*}{6{,}000}
 & Risk & $-0.21 \pm 0.00$ & $0.77 \pm 0.00$ & $1.18 \pm 0.01$ & --- \\
 & DR-Emb & $-0.52 \pm 0.01$ & $0.66 \pm 0.00$ & $1.12 \pm 0.00$ & $0.57 \pm 0.01$ \\
 & DragonNet & $-0.45 \pm 0.01$ & $0.68 \pm 0.00$ & $1.14 \pm 0.00$ & $0.65 \pm 0.01$ \\
 & TARNet & $-0.52 \pm 0.02$ & $0.66 \pm 0.01$ & $1.13 \pm 0.00$ & $0.57 \pm 0.03$ \\
 & AACE (ours) & $\mathbf{-0.57 \pm 0.01}$ & $\mathbf{0.64 \pm 0.00}$ & $\mathbf{1.11 \pm 0.00}$ & $\mathbf{0.52 \pm 0.01}$ \\
\midrule
\multirow{5}{*}{8{,}000}
 & Risk & $-0.23 \pm 0.00$ & $0.77 \pm 0.00$ & $1.18 \pm 0.01$ & --- \\
 & DR-Emb & $-0.53 \pm 0.01$ & $0.65 \pm 0.00$ & $1.12 \pm 0.00$ & $0.56 \pm 0.01$ \\
 & DragonNet & $-0.48 \pm 0.01$ & $0.67 \pm 0.00$ & $1.13 \pm 0.00$ & $0.63 \pm 0.02$ \\
 & TARNet & $-0.53 \pm 0.00$ & $0.65 \pm 0.00$ & $1.12 \pm 0.00$ & $0.58 \pm 0.02$ \\
 & AACE (ours) & $\mathbf{-0.58 \pm 0.00}$ & $\mathbf{0.64 \pm 0.00}$ & $\mathbf{1.11 \pm 0.00}$ & $\mathbf{0.50 \pm 0.01}$ \\
\midrule
\multicolumn{2}{l}{Oracle} & $-0.74$ & $0.61$ & $1.08$ & 0.00 \\
\bottomrule
\end{tabular}
\end{footnotesize}
\end{table*}

\paragraph{Extended Robustness Analysis.}
We complement the robustness analyses in the main text by reporting additional metrics beyond AUTOC. We first report extended results for noisy annotations, missing annotated confounders, and added irrelevant annotation variables (Tables~\ref{tab:synsum_annotation_corruption}--\ref{tab:synsum_irrelevant_annotations}). We then report extended results for representation masking and the alternative encoder experiment (Tables~\ref{tab:synsum_masking} and~\ref{tab:synsum_gemma}). Overall, these results support the main finding that \textsc{AACE} remains effective under moderate degradation of annotation and representation quality.

\begin{table*}[h]
\centering
\caption{\textbf{\textsc{SynSum} (noisy annotations).} Policy metrics of \textsc{AACE} as a function of annotation corruption, using 4{,}000 training samples and a fixed test set (mean~$\pm$~std.~dev. over five random seeds). The first column indicates the percentage of confounder annotations corrupted before training. Lower values are better.}
\label{tab:synsum_annotation_corruption}
\renewcommand{\arraystretch}{1}
\begin{footnotesize}
\begin{tabular}{ccccc}
\toprule
\textbf{Corruption (\%)} & \textbf{AUTOC $(\downarrow)$} & \textbf{$\boldsymbol{\overline{V}}$ $(\downarrow)$} & \textbf{$\boldsymbol{V_{10\%}}$ $(\downarrow)$} & \textbf{PEHE $(\downarrow)$} \\
\midrule
0   & $-0.56 \pm 0.01$ & $0.65 \pm 0.00$ & $1.11 \pm 0.01$ & $0.53 \pm 0.02$ \\
5   & $-0.55 \pm 0.01$ & $0.65 \pm 0.00$ & $1.11 \pm 0.00$ & $0.55 \pm 0.01$ \\
10  & $-0.53 \pm 0.01$ & $0.66 \pm 0.00$ & $1.11 \pm 0.00$ & $0.58 \pm 0.02$ \\
25  & $-0.49 \pm 0.01$ & $0.67 \pm 0.01$ & $1.12 \pm 0.00$ & $0.64 \pm 0.02$ \\
50  & $-0.42 \pm 0.04$ & $0.69 \pm 0.02$ & $1.15 \pm 0.01$ & $0.78 \pm 0.05$ \\
75  & $-0.39 \pm 0.03$ & $0.70 \pm 0.01$ & $1.15 \pm 0.01$ & $0.85 \pm 0.06$ \\
100 & $-0.37 \pm 0.06$ & $0.70 \pm 0.01$ & $1.16 \pm 0.02$ & $0.89 \pm 0.06$ \\
\bottomrule
\end{tabular}
\end{footnotesize}
\end{table*}

\begin{table*}[h]
\centering
\caption{\textbf{\textsc{SynSum} (missing relevant confounders).} Policy metrics of \textsc{AACE} as a function of the number of removed annotated confounders. See Table~\ref{tab:synsum_annotation_corruption} for details. Lower values are better.}
\label{tab:synsum_dropped_annotations}
\renewcommand{\arraystretch}{1}
\begin{footnotesize}
\begin{tabular}{ccccc}
\toprule
\textbf{Dropped (n)} & \textbf{AUTOC $(\downarrow)$} & \textbf{$\boldsymbol{\overline{V}}$ $(\downarrow)$} & \textbf{$\boldsymbol{V_{10\%}}$ $(\downarrow)$} & \textbf{PEHE $(\downarrow)$} \\
\midrule
0 & $-0.56 \pm 0.01$ & $0.65 \pm 0.00$ & $1.11 \pm 0.01$ & $0.53 \pm 0.02$ \\
1 & $-0.51 \pm 0.03$ & $0.67 \pm 0.02$ & $1.12 \pm 0.00$ & $0.60 \pm 0.05$ \\
2 & $-0.50 \pm 0.01$ & $0.67 \pm 0.01$ & $1.12 \pm 0.01$ & $0.61 \pm 0.02$ \\
3 & $-0.47 \pm 0.04$ & $0.68 \pm 0.01$ & $1.13 \pm 0.01$ & $0.68 \pm 0.07$ \\
4 & $-0.41 \pm 0.03$ & $0.69 \pm 0.01$ & $1.14 \pm 0.01$ & $0.77 \pm 0.03$ \\
5 & $-0.38 \pm 0.05$ & $0.70 \pm 0.01$ & $1.16 \pm 0.01$ & $0.88 \pm 0.05$ \\
\bottomrule
\end{tabular}
\end{footnotesize}
\end{table*}

\begin{table*}[h]
\centering
\caption{\textbf{\textsc{SynSum} (irrelevant annotations).} Policy metrics of \textsc{AACE} as a function of the number of added irrelevant annotation variables. See Table~\ref{tab:synsum_annotation_corruption} for details. Lower values are better.}
\label{tab:synsum_irrelevant_annotations}
\renewcommand{\arraystretch}{1}
\begin{footnotesize}
\begin{tabular}{ccccc}
\toprule
\textbf{Added (n)} & \textbf{AUTOC $(\downarrow)$} & \textbf{$\boldsymbol{\overline{V}}$ $(\downarrow)$} & \textbf{$\boldsymbol{V_{10\%}}$ $(\downarrow)$} & \textbf{PEHE $(\downarrow)$} \\
\midrule
0 & $-0.56 \pm 0.01$ & $0.65 \pm 0.00$ & $1.11 \pm 0.01$ & $0.53 \pm 0.02$ \\
1 & $-0.56 \pm 0.01$ & $0.65 \pm 0.00$ & $1.11 \pm 0.00$ & $0.53 \pm 0.01$ \\
2 & $-0.56 \pm 0.01$ & $0.65 \pm 0.00$ & $1.11 \pm 0.00$ & $0.53 \pm 0.01$ \\
3 & $-0.56 \pm 0.01$ & $0.65 \pm 0.00$ & $1.11 \pm 0.00$ & $0.54 \pm 0.01$ \\
4 & $-0.56 \pm 0.01$ & $0.65 \pm 0.00$ & $1.11 \pm 0.01$ & $0.53 \pm 0.01$ \\
5 & $-0.55 \pm 0.01$ & $0.65 \pm 0.00$ & $1.11 \pm 0.01$ & $0.54 \pm 0.01$ \\
\bottomrule
\end{tabular}
\end{footnotesize}
\end{table*}

\begin{table*}[h]
\centering
\caption{\textbf{\textsc{SynSum} (representation masking).} Policy metrics under random masking of embedding dimensions, using 4{,}000 training samples and a fixed test set (mean~$\pm$~std.~dev. over five random seeds). The first column indicates the percentage of embedding dimensions randomly masked for all methods. Lower values are better. PEHE measures treatment effect estimation error and is not defined for the risk-based model. The best model within each masking level is shown in \textbf{bold}. The \textit{oracle} row reports the metrics of a policy that ranks by the true treatment effect.}
\label{tab:synsum_masking}
\renewcommand{\arraystretch}{1}
\begin{footnotesize}
\begin{tabular}{lccccc}
\toprule
\textbf{Masking (\%)} & \textbf{Model} & \textbf{AUTOC $(\downarrow)$} & \textbf{$\boldsymbol{\overline{V}}$ $(\downarrow)$} & \textbf{$\boldsymbol{V_{10\%}}$ $(\downarrow)$} & \textbf{PEHE $(\downarrow)$} \\
\midrule

\multirow{5}{*}{0}
 & Risk & $-0.22 \pm 0.01$ & $0.77 \pm 0.00$ & $1.18 \pm 0.00$ & --- \\
 & DR-Emb & $-0.51 \pm 0.01$ & $0.66 \pm 0.00$ & $1.12 \pm 0.00$ & $0.59 \pm 0.02$ \\
 & DragonNet & $-0.40 \pm 0.02$ & $0.70 \pm 0.00$ & $1.14 \pm 0.01$ & $0.71 \pm 0.02$ \\
 & TARNet & $-0.51 \pm 0.02$ & $0.66 \pm 0.00$ & $1.13 \pm 0.01$ & $0.58 \pm 0.02$ \\
 & AACE (ours) & $\mathbf{-0.56 \pm 0.01}$ & $\mathbf{0.65 \pm 0.00}$ & $\mathbf{1.11 \pm 0.01}$ & $\mathbf{0.53 \pm 0.02}$ \\
\midrule

\multirow{5}{*}{75}
 & Risk & $-0.21 \pm 0.01$ & $0.77 \pm 0.00$ & $1.19 \pm 0.00$ & --- \\
 & DR-Emb & $-0.47 \pm 0.01$ & $0.68 \pm 0.00$ & $1.13 \pm 0.01$ & $0.64 \pm 0.02$ \\
 & DragonNet & $-0.28 \pm 0.05$ & $0.74 \pm 0.01$ & $1.17 \pm 0.02$ & $0.80 \pm 0.03$ \\
 & TARNet & $-0.46 \pm 0.04$ & $0.67 \pm 0.01$ & $1.14 \pm 0.01$ & $0.65 \pm 0.06$ \\
 & AACE (ours) & $\mathbf{-0.54 \pm 0.01}$ & $\mathbf{0.66 \pm 0.00}$ & $\mathbf{1.12 \pm 0.00}$ & $\mathbf{0.56 \pm 0.01}$ \\
\midrule

\multirow{5}{*}{90}
 & Risk & $-0.19 \pm 0.01$ & $0.78 \pm 0.00$ & $1.20 \pm 0.00$ & --- \\
 & DR-Emb & $-0.43 \pm 0.01$ & $0.69 \pm 0.00$ & $1.13 \pm 0.01$ & $0.68 \pm 0.02$ \\
 & DragonNet & $-0.27 \pm 0.04$ & $0.75 \pm 0.01$ & $1.17 \pm 0.02$ & $0.86 \pm 0.02$ \\
 & TARNet & $-0.44 \pm 0.03$ & $0.68 \pm 0.01$ & $1.14 \pm 0.01$ & $0.67 \pm 0.06$ \\
 & AACE (ours) & $\mathbf{-0.51 \pm 0.01}$ & $\mathbf{0.67 \pm 0.01}$ & $\mathbf{1.12 \pm 0.00}$ & $\mathbf{0.60 \pm 0.01}$ \\
\midrule

\multirow{5}{*}{95}
 & Risk & $-0.17 \pm 0.02$ & $0.78 \pm 0.00$ & $1.20 \pm 0.00$ & --- \\
 & DR-Emb & $-0.36 \pm 0.03$ & $0.71 \pm 0.00$ & $1.16 \pm 0.01$ & $0.74 \pm 0.01$ \\
 & DragonNet & $-0.29 \pm 0.02$ & $0.75 \pm 0.01$ & $1.15 \pm 0.00$ & $0.89 \pm 0.03$ \\
 & TARNet & $-0.39 \pm 0.03$ & $0.70 \pm 0.01$ & $1.15 \pm 0.01$ & $0.73 \pm 0.02$ \\
 & AACE (ours) & $\mathbf{-0.46 \pm 0.00}$ & $\mathbf{0.69 \pm 0.00}$ & $\mathbf{1.13 \pm 0.01}$ & $\mathbf{0.65 \pm 0.01}$ \\
\midrule

\multirow{5}{*}{100}
 & Risk & $-0.18 \pm 0.01$ & $0.78 \pm 0.00$ & $1.20 \pm 0.00$ & --- \\
 & DR-Emb & $-0.28 \pm 0.02$ & $0.76 \pm 0.01$ & $\mathbf{1.15 \pm 0.00}$ & $0.93 \pm 0.01$ \\
 & DragonNet & $-0.27 \pm 0.03$ & $0.76 \pm 0.01$ & $1.16 \pm 0.01$ & $0.95 \pm 0.02$ \\
 & TARNet & $-0.27 \pm 0.01$ & $0.76 \pm 0.00$ & $1.15 \pm 0.01$ & $0.94 \pm 0.03$ \\
 & AACE (ours) & $\mathbf{-0.32 \pm 0.01}$ & $\mathbf{0.74 \pm 0.00}$ & $1.15 \pm 0.01$ & $\mathbf{0.75 \pm 0.00}$ \\
\midrule

\multicolumn{2}{l}{Oracle} & $-0.74$ & $0.61$ & $1.08$ & $0.00$ \\

\bottomrule
\end{tabular}
\end{footnotesize}
\end{table*}

\clearpage

\begin{table*}[h]
\centering
\caption{\textbf{\textsc{SynSum} (alternative encoder).} Policy metrics of all methods using Gemma embeddings, using 4{,}000 training samples and a fixed test set (mean~$\pm$~std.~dev. over five random seeds). Lower values are better. PEHE measures treatment effect estimation error and is not defined for the risk-based model. The best model is shown in \textbf{bold}.}
\label{tab:synsum_gemma}
\renewcommand{\arraystretch}{1}
\begin{footnotesize}
\begin{tabular}{lcccc}
\toprule
\textbf{Model} & \textbf{AUTOC $(\downarrow)$} & \textbf{$\boldsymbol{\overline{V}}$ $(\downarrow)$} & \textbf{$\boldsymbol{V_{10\%}}$ $(\downarrow)$} & \textbf{PEHE $(\downarrow)$} \\
\midrule
Risk & $-0.25 \pm 0.01$ & $0.77 \pm 0.00$ & $1.17 \pm 0.00$ & --- \\
DR-Emb & $-0.57 \pm 0.01$ & $0.65 \pm 0.00$ & $1.11 \pm 0.00$ & $0.53 \pm 0.02$ \\
DragonNet & $-0.48 \pm 0.00$ & $0.68 \pm 0.00$ & $1.13 \pm 0.00$ & $0.63 \pm 0.01$ \\
TARNet & $-0.54 \pm 0.03$ & $0.66 \pm 0.01$ & $1.12 \pm 0.00$ & $0.56 \pm 0.04$ \\
AACE (ours) & $\mathbf{-0.60 \pm 0.00}$ & $\mathbf{0.64 \pm 0.00}$ & $\mathbf{1.10 \pm 0.00}$ & $\mathbf{0.48 \pm 0.02}$ \\
\bottomrule
\end{tabular}
\end{footnotesize}
\end{table*}

\subsection{MIMIC-Syn}
\label{app:mimicsyn_results}

\paragraph{Recoverability of Confounders From Text.}
As in \textsc{SynSum}, we assess how well the confounders used in the data-generating process can be recovered from the text representations. Specifically, we train classifiers to predict the cardiovascular diagnosis variables from the clinical notes. As shown in Table~\ref{tab:mimic_confounders}, these confounders are predictable from text, although imperfectly, indicating that the notes contain substantial confounding signal.

\begin{table*}[h]
\centering
\begin{footnotesize}
\caption{\textbf{\textsc{MIMIC-Syn} (confounder predictability from text).} Prediction of diagnostic confounders from clinical notes (mean $\pm$ std. over five random seeds). ``Prev.'' denotes the positive-class prevalence in the test set. Each classifier's decision threshold was selected to maximize F1 score on the validation set.}
\renewcommand{\arraystretch}{1}
\label{tab:mimic_confounders}
\begin{tabular}{lcccccc}
\toprule
\textbf{Confounder} & \textbf{Prev.} & \textbf{F1} & \textbf{Acc.} & \textbf{AUPRC} & \textbf{AUROC} \\
\midrule
Atrial fibrillation   & 0.24 & $0.694 \pm 0.007$ & $0.856 \pm 0.005$ & $0.759 \pm 0.004$ & $0.889 \pm 0.002$ \\
Congestive heart failure         & 0.22 & $0.591 \pm 0.005$ & $0.794 \pm 0.009$ & $0.584 \pm 0.004$ & $0.845 \pm 0.002$ \\
Coronary atherosclerosis & 0.26 & $0.766 \pm 0.003$ & $0.884 \pm 0.006$ & $0.854 \pm 0.004$ & $0.917 \pm 0.002$ \\
Hypertension           & 0.44 & $0.648 \pm 0.002$ & $0.573 \pm 0.009$ & $0.615 \pm 0.002$ & $0.694 \pm 0.002$ \\
\bottomrule
\end{tabular}
\end{footnotesize}
\end{table*}

\subsection{\textsc{MIMIC-Real}}
\label{app:mimicreal_results}

\paragraph{Additional Analysis of Treatment Assignments.}
Since ground-truth treatment effects are unavailable in \textsc{MIMIC-Real}, we cannot determine which method produces the best policy. We therefore restrict the analysis to agreement between the treatment rankings induced by the different methods, and between \textsc{AACE} and the observed treatments. Table~\ref{tab:real_policy_full} reports the corresponding Jaccard overlaps and Spearman rank correlations. The results show that \textsc{AACE} is more strongly aligned with the representation-based causal baseline than with the risk-based model, especially in terms of global rank correlation. At the same time, agreement between \textsc{AACE} and the observed treatment assignments is low, indicating that the learned rankings differ substantially from the treatment decisions recorded in practice. Overall, these results reinforce the main paper’s qualitative conclusion that annotation-assisted, risk-based, and representation-based causal strategies can prioritize different patient groups on real-world EHR data.

\begin{table*}[h]
\centering
\caption{\textbf{\textsc{MIMIC-Real} (policy overlap and agreement).} Policy overlap and ranking agreement on real-world data. Jaccard measures agreement between top-$k$ treatment assignments, while Spearman $\rho$ captures global rank correlation. The final column compares the learned ranking of \textsc{AACE} with observed treatment assignment and is reported separately because observed treatment is binary rather than a full ranking.}
\label{tab:real_policy_full}
\renewcommand{\arraystretch}{1}
\begin{footnotesize}
\begin{tabular}{lccc|c}
\toprule
\textbf{Metric} & \textbf{AACE vs TARNet} & \textbf{AACE vs Risk} & \textbf{Risk vs TARNet} & \textbf{AACE vs Observed} \\
\midrule
\multicolumn{5}{c}{Jaccard overlap} \\
\midrule
$k = 5\%$  & $0.088 \pm 0.106$ & $0.037 \pm 0.027$ & $0.316 \pm 0.166$ & $0.034 \pm 0.004$ \\
$k = 10\%$ & $0.155 \pm 0.115$ & $0.076 \pm 0.048$ & $0.321 \pm 0.170$ & $0.055 \pm 0.006$ \\
$k = 20\%$ & $0.245 \pm 0.088$ & $0.141 \pm 0.052$ & $0.340 \pm 0.148$ & $0.078 \pm 0.022$ \\
$k = 50\%$ & $0.573 \pm 0.075$ & $0.362 \pm 0.051$ & $0.421 \pm 0.126$ & $0.146 \pm 0.014$ \\
\midrule
\multicolumn{5}{c}{Spearman correlation} \\
\midrule
$\rho$ & $0.551 \pm 0.202$ & $0.067 \pm 0.170$ & $0.233 \pm 0.313$ & $0.070 \pm 0.057$ \\
\bottomrule
\end{tabular}
\end{footnotesize}
\end{table*}

\clearpage
\section{Experimental Details}
\label{app:details}
This section summarizes implementation details common to all experiments. We use the same architecture, tuning procedure, and early stopping strategy across datasets and model components. All models (classifiers, nuisance models, and treatment effect regressors) were implemented as small neural networks with a single hidden layer and ReLU activations, using a sigmoid output layer for binary classification tasks and a linear output layer for regression tasks. Models were trained with Adam, optional weight decay, and a maximum of 50~epochs, with early stopping based on validation loss (patience of five epochs). The confounder recoverability experiments used class-weighted binary cross-entropy, with positive class weight $(1-p)/p$, where $p$ is the positive-class prevalence in the training set, other classification tasks used standard binary cross-entropy, and regression tasks used mean squared error. For the doubly robust pseudo outcomes, we do not apply cross-fitting. Hyperparameters were tuned once per model family on a single training size and then reused across experiments. We performed a random search with 12~trials over the following grid: hidden dimension $\in \{64,128\}$, learning rate $\in \{10^{-4}, 3\times10^{-4}, 5\times10^{-4}, 10^{-3}\}$, weight decay $\in \{0,10^{-5},10^{-4}\}$, and batch size $\in \{128,256\}$. The configuration with the lowest validation loss was selected. All experiments used 10\% of the data for validation and 10\% for testing, with the remaining data used for training. Random seeds affected both data splits and weight initialization.

\end{document}